\documentclass[12pt]{article}

\usepackage{amsmath}

\usepackage{newtxtext,newtxmath}

\usepackage{graphicx}

\usepackage[letterpaper,margin=1in]{geometry}

\renewenvironment{abstract}
	{\quotation}
	{\endquotation}

\date{}

\makeatletter
\renewcommand{\fnum@figure}{\textbf{Figure \thefigure}}
\renewcommand{\fnum@table}{\textbf{Table \thetable}}
\makeatother

\usepackage{scicite}

\usepackage[hyphens]{url}
\makeatletter\g@addto@macro{\UrlBreaks}{\UrlOrds}\makeatother
\usepackage{scrextend}
\usepackage{float}
\usepackage{color}
\usepackage{caption}
\usepackage{subcaption}

\newcommand{\dc}[3]{d_{c_{#1}}\!\left(#2,#3\right)}

\newtheorem{theorem}{Theorem}
\newtheorem{corollary}{Corollary}
\newtheorem{remark}{Remark}
\newtheorem{lemma}{Lemma}

\newtheorem{assumption}{Assumption}
\newtheorem{definition}{Definition}
\newtheorem{proposition}{Proposition}

\def\scititle{
	A general-purpose safe manipulator planning framework for diverse environments without retraining
}
\title{\bfseries \boldmath \scititle}

\author{
	Mingkang~Long$^{1}$, Rong~Su$^{1\ast}$, Luhang~Hong$^{1}$, Qingqing~Wang$^{1}$,
	Ruochen~Tai$^{2}$\and
	\small$^{1}$Nanyang Technological University\and
	\small$^{2}$Huazhong University of Science and Technology\and
	\small$^\ast$Corresponding author. Email: rsu@ntu.edu.sg
}

\begin{document}

\maketitle

\begin{abstract} \bfseries \boldmath
Nowadays, robots need to be retrained whenever their operating environment changes, and this prohibitive training cost keeps them far from daily use. The community is calling for innovations that bestow on a robot the ability to safely grasp different objects under different conditions without excessive training data and computing power. In response to this call, we present a general-purpose manipulator mission-planning framework, allowing a robot to safely perform manipulation tasks, such as grasping objects that fit in its gripper, across different operating environments without retraining its controllers. The framework rests on two innovations: a {\em finite-state abstracted model}, bisimilar to the original system dynamics, that describes accessibility of the space in front of the robot under its given arm and gripper, and a {\em block-conservative specification generation method}  that turns each manipulation task under any specific operational environment into specifications about state accessibility of the abstracted model. A task proven feasible and safe on the abstracted model is guaranteed to be safe and achievable by the robot. Real-world case studies conducted on a Unitree G1 U1 humanoid show that manipulation tasks commonly encountered in daily life, such as grasping an object, pressing an elevator button or picking up a berry, can be handled within this framework, where an abstracted model of the G1 U1 robot is built only once to avoid costly model reconstructions, and different tasks are simply captured by specific block-conservative specifications, achievable by a generic mission planner with vision-based feedback control.
\end{abstract}

\section*{Introduction}
With the advancement of artificial intelligence (AI), in particular, via the development of foundation models and reinforcement learning algorithms, robots have become more intelligent, capable of handling a variety of tasks. However, the inherent prohibitively high training cost in building meaningful foundation models or tuning parameters of a learned model makes a robot far from being a necessity in our daily life. How to build a general-purpose robot that can handle various tasks under different operational environments without incurring high (re-)training costs is a long-standing challenge. It is certainly not our intention to propose a general solution to that challenge in this paper. However, we do want to argue with theoretical and practical evidence that, for a commonly used robotic function about object grasping, such a general-purpose solution may exist. More explicitly, for a given robot equipped with an arm-motion controller and a gripper controller, a general-purpose mission planner exists, which, by leveraging these two controllers, can grasp any object fitting in the gripper, e.g., an apple in a tree, a workpiece inside a machine or a cola can on a table, without retraining. To derive a general-purpose mission planner for object grasping, we need two ingredients: a model that describes what a robot can do, and specifications that describe what the robot should do, leveraging the derived model.

There is a rich volume of research on modeling of robotic requirements upon formal models of robotic dynamics. For example, automata-based~\cite{belta2016formal} and symbolic-control~%
\cite{pola2019control} methods deal with finite-state systems
whose behavioral requirements are described by linear temporal logic (LTL). To handle the continuous state space of a real robot, which is infeasible for model checking and formal synthesis, a \textit{finite-state model
abstraction} is needed~\cite{girard2011approximate}. Creating an effective abstraction of continuous system dynamics
has become a cornerstone of correct-by-design synthesis for safety-critical
autonomous systems~\cite{yin2024formal}. Various finite abstraction methods have been proposed in the literature, from exact constructions for structured systems~\cite{alur1994theory,henzinger1995s,tabuada2003model,lafferriere1999new} and
classical quantization- or invariant-based techniques~\cite{moor2002discrete,%
bicchi2002reachability,forstner2002discrete,koutsoukos2000supervisory,%
caines1998hierarchical} to \textit{approximately bisimilar} models, which trade exact
equivalence for a bounded, certifiable error~\cite{girard2007approximation,%
girard2011approximate} and cover linear, switched, nonlinear, distributed, and
stochastic dynamics~\cite{Girard2007Approximately,girard2009approximately,%
zamani2012symbolic,meyer2018compositional,mizoguchi2018deadlock,%
pola2008approximately,lin2024bisimulation}. Data-driven learning
methods can infer abstractions or certificates directly from data~\cite{jackson2020safety,%
chen2021temporal,hashimoto2022learning,nejati2023formal,edwards2025general,%
abate2024bisimulation}, including Koopman liftings of nonlinear
dynamics~\cite{bevanda2021koopman,sinha2022data,lusch2018deep}.

In those existing works, an abstracted model is derived from both the original continuous dynamics and a target
 set of task-specific requirements (and, for learned models, from
the training data). Any change in task requirements may force the model to be
regenerated, retrained, or re-synthesized, which is costly and lacks real-time responsiveness. For a robot expected to perform many tasks
over its lifetime, this per-task rebuild
is a central bottleneck to \textit{certified} autonomy. Symbolic-control
frameworks based on feedback-refinement relations attempt to loosen the coupling by
building an abstracted model from the underlying dynamics and a static quantizer alone, upon which
the specifications are defined over states of the abstracted model afterwards~\cite{reissig2017feedback,rungger2016scots}. However, such methods re-quantize the state space at every step, so that the states of the original dynamics mapped into the same abstract state are in general not bisimilar; indeed, a quotient of a continuous system is bisimilar to it only under structural conditions on the dynamics~\cite{pappas2003bisimilar,%
tabuada2004bisimilar}. The consequence is that, first of all, the abstracted model is nondeterministic and a controller synthesized on it transfers to the original dynamics only with a conservatism that grows with the quantization error, and secondly, the construction of the abstracted model is more computationally demanding.

In this paper, by introducing a new distance between two states under linear dynamics, we show that the continuous state space is uniquely partitioned into a Voronoi diagram with respect to a properly chosen set of seeds. The next set of seeds is uniquely determined by the current set  and a given control signal under the linear dynamics, which induces another unique Voronoi diagram. By choosing an initial set of seeds, our abstracted model is essentially a tree of Voronoi diagrams, where each edge corresponds to a specific control signal that links the seed sets of two neighboring Voronoi diagrams. We can show that, within this (infinite) tree, any two states in the continuous space mapped to the same initial state (or Voronoi cell) are bisimilar to each other with respect to any sequence of control signals permitted in the tree. When an infinite planning horizon is considered, the derived tree will become infinite, making formal synthesis infeasible. Fortunately, in this case, if the target system dynamics is asymptotically stable (or is made so by an inner feedback loop), the infinite tree can be cut short, leading to an approximate bisimulation relation over trajectories in the continuous space. The resulting modeling inaccuracy is upper bounded by an analytic function over the initial setting. Upon such an abstracted model, we present an innovative method to transform each grasping task into a set of formal block-conservative requirements over the states of the abstracted model, and show that any grasping task that can be safely solved over the abstracted model must render a safe grasping control sequence in the original system, as long as a target object fits in the gripper. The validity and effectiveness of this new framework (Fig.~\ref{fig:pipeline}) are illustrated both in
simulation and on a real Unitree G1 U1 humanoid robot, where the same arm-motion and gripper controllers are applied in different operational cases without any retraining for model reconstruction. Each formally derived grasping plan shown to be safe with respect to the abstracted model renders safe object grasping in the real world. Formal statements are collected in Materials and Methods and proofs in the Supplementary Text.

\section*{Results}

\subsection*{Problem statement of robotic manipulator control}
A modern robotic manipulator is delivered with a vendor-supplied low-level control
stack: an embedded servo loop performing torque
control, gravity compensation and feedback linearization; an inverse-kinematics solver;
an \textit{arm-stretching} (reaching) function that drives the arm to a commanded joint
configuration or end-effector pose; and a \textit{finger} function that opens and closes
the gripper. These functions are engineered and validated by the vendor, and each is
reliable at the level of a single motion. What the vendor stack does not supply is a
guarantee at the level of \textit{mission planning}. Whether the particular sequence of reaches
and grasps that accomplishes a task will keep the entire arm clear of the obstacles that
happen to be present, respect the rules of the workspace, and terminate in the goal
configuration is not decided by any of these primitives individually; it is a property of
their composition, and it changes every time the environment or the objective changes.

This work addresses that mission planning layer. We treat the platform's
existing functions as given primitives and do not modify, retrain, or replace them: servo
control, inverse kinematics, arm stretching, finger opening and closing, whole-body
balance and perception are all consumed as they are shipped. Such a treatment spares us the typically costly retraining of basic primitives. Our main contribution is to propose a certifiable and multi-purpose
\textit{mission-planning} layer that sits above those primitives, that is, to decide \textit{which}
sequence of primitive invocations to issue, and to certify \textit{before} any motion is
executed that the resulting whole-arm trajectory satisfies any given mission specifications. 

The problem that we solve is described as follows. Consider a manipulator whose user-level joint-space
behavior under the vendor's position interface is a discrete-time dynamic model
(Eq.~\ref{cobot_dynamic}), driven by a finite set $U$ of admissible joint-space increments,
and a mission specification $\phi$ expressed in a chosen (not necessarily linear) temporal logic over regions of the
configuration and workspace, e.g., obstacles to be avoided, targets to be reached, orderings to be
respected. We ask for an abstracted model of the manipulator that (i)~is constructed
from the given dynamics alone, independent of $\phi$, via quotient construction with respect to the bisimilarity relation over dynamic behaviors, so
that a property verified on the model holds on the physical arm; and (ii)~accommodates a new
mission by \textit{updating} temporal properties over each existing cell within the abstracted model rather than by rebuilding the abstracted model. Given such a
model, we ask further to decide whether a candidate mission plan satisfies $\phi$, and, when
it does not, to synthesize one that does, returned as a sequence of the platform's existing
arm-stretching and finger closing commands. The remainder of the Results establishes each part in
turn.

\subsection*{Construction of multi-purpose abstracted models}
The model is \textit{multi-purpose} in a precise sense: for a given robot, one abstracted
model, built once from that robot's own dynamics and never rebuilt, serves every mission
that robot is asked to perform (Fig.~\ref{fig:concept}). Across the missions reported
below only the (temporal logic) property labels attached to existing cells changed; the model, the verification
procedure, and the synthesis procedure were identical. The manipulator's abstraction, a
joint-space grid with a precomputed per-voxel collision map, is built once offline.
Thereafter, each new mission is encoded by property relabeling alone: the reachable set is
recomputed for its obstacles with the collision map resident in memory, and reachability-analysis-based synthesis
returns a certified path on the property-relabeled grid. The per-mission cost is thus confined to
relabeling and search.

The three guarantees behind this behavior are proved in Materials and Methods (proofs
in the Supplementary Text). The state space is partitioned along the eigen-directions of
the system matrix: for a finite and eigen-separable set of representative points (Definition~\ref{definition2}), the neighborhood cells defined in Materials and Methods form a Voronoi diagram $\mathcal{V}$ with respect to the eigen-coordinate distances, whose seeds are the representative points. From now on, we use $v\in\mathcal{V}$ both for a cell of $\mathcal{V}$ and for its seed, and write $N(v)$ for the cell when the distinction matters. Under the dynamics (Eq.~\ref{eq01}), each control action $u\in U$ maps every seed of $\mathcal{V}$ to a new state. The collection of these new states is again eigen-separable and induces a unique Voronoi diagram $\mathcal{V}'$, called the {\em offspring} of $\mathcal{V}$ with respect to $u$. An abstracted model is a transition system $T$, not necessarily finite, whose state set $V$ is the union of the cells of the Voronoi diagrams computed iteratively as follows: (1) choose an initial eigen-separable seed set, which induces an initial Voronoi diagram $\mathcal{V}_0$, and (2) iteratively derive the offspring of every Voronoi diagram generated in the previous iteration with respect to each control action. A transition $(v,u,v')$ is valid in this abstracted model if $v$ and $v'$ are cells of two Voronoi diagrams $\mathcal{V}$ and $\mathcal{V}'$, respectively, such that $\mathcal{V}'$ is the offspring of $\mathcal{V}$ with respect to $u$ and the seed of $v'$ is the image of the seed of $v$ under (Eq.~\ref{eq01}) with input $u$. In this abstracted model $T$, if two states of the original discrete-time dynamics fall in the same state of $T$, i.e., in the same cell of a Voronoi diagram, then they are bisimilar with respect to all 
control sequences (Theorem~\ref{theorem1}). That is, under each control sequence and the dynamics (Eq.~\ref{eq01}), their target states must fall in the same cell of the resulting Voronoi diagram.  Each cell size is bounded explicitly
(Lemma~\ref{lemma1} and Lemma~\ref{lemma_celldiam}). Due to this bisimulation result, for any (block conservative) temporal-logic specification defined over the states of $T$, if it holds on the abstracted model with respect to a set of control sequences, then it must hold for each state on the original
system (Theorem~\ref{theorem2}) under the same set of control sequences. If the given robotic system is asymptotically stable, a finite abstracted model can be built that ensures approximate bisimulation of any two states falling in the same cell. In this case, the maximum distance between two trajectories originating from these two states is bounded in terms of the size of the corresponding seed set and the maximum length of control sequences to be considered (Theorem~\ref{theorem_approx} and Theorem~\ref{theorem3}). To illustrate this framework for a manipulator, we present a specific discrete-time dynamic
model (Eq.~\ref{cobot_dynamic}), commonly seen in reach--avoid mission environments, whose system matrix $A$ is the identity $I$. By choosing a uniformly distributed seed set, the corresponding initial Voronoi diagram is a uniform joint-space grid, and, over the finite planning horizon that reach--avoid missions require, the abstracted model $T$ is finite and exactly bisimilar.

\paragraph*{Bisimulation relation is preserved under the dynamics}
We first confirm that, for any two cells $v$ and $\hat{v}$ of a Voronoi diagram $\mathcal{V}$ under any control action $u$, their common boundary, i.e., the set of states at equal eigen-coordinate distance from the two seeds, is mapped onto the common boundary of two cells $v'$ and $\hat{v}'$ of the offspring $\mathcal{V}'$ whose seeds are the images of the seeds of $v$ and $\hat{v}$ under Eq.~\ref{eq01} with input $u$ (Theorem~\ref{theorem1}). As an illustration, consider
Eq.~\ref{eq01} with
\begin{equation*}
A=
\begin{bmatrix}
1 & -0.5\\
0.4 & 1
\end{bmatrix},
\qquad
B=
\begin{bmatrix}
1 & 0\\
0 & 1
\end{bmatrix}.
\end{equation*}
One left eigenvector of $A$ is $c^H=[\,0.6667\mathrm{i}\ \ 0.7454\,]$. Here $n=2$ with
a complex eigenvalue, so a single eigen-direction already separates the plane
($r=1$, $d_1=2$) and Assumption~\ref{assumption_diag} holds. Since
we consider bisimulation of Theorem~\ref{theorem1} over an infinite horizon, no stability requirement is
imposed. The eigen-coordinate matrix of Eq.~\ref{eq_eigencoord} is
$\Pi=\left[\begin{smallmatrix}0&0.7454\\0.6667&0\end{smallmatrix}\right]$, whose singular values $0.6667$ and $0.7454$ give
$\underline{\alpha}=0.6667$ and $\operatorname{cond}(\Pi)=1.118$, so the eigen-coordinate map is well conditioned. The seed set is chosen as \[\mathcal{I}_A=\{[-1,-2]^T,[-2,-3]^T,[-3,-1]^T,[2,1]^T,[-1,2]^T,[2,3]^T,[3,1]^T\}.\] Since any finite set is eigen-separable when $r=1$, by Lemma~\ref{lemma1} every state lies within
$\sqrt{r}\,\varepsilon/\underline{\alpha}=1.50\,\varepsilon$ of the closest seed point in $\mathcal{I}_A$, where $\varepsilon$ is the covering radius of $\mathcal{I}_A$.
Figure~\ref{fig1}A shows the Voronoi diagram and the chosen seeds, together with
randomly sampled states colored by the cell that contains them. Applying the common input $u=[1,0]^T$ to every
seed, Fig.~\ref{fig1}D shows the offspring Voronoi diagram after one transition. Any two states within the same cell of the Voronoi diagram in Fig.~\ref{fig1}A will be mapped to two new states within the same cell of the offspring Voronoi diagram in Fig.~\ref{fig1}D; the sampled states in Fig.~\ref{fig1}, color-coded by their initial cell, confirm this behavior, as no color crosses a cell boundary after the transition. If we define two states to be equivalent if they stay in the same cell of a Voronoi diagram, then any propagated states from two equivalent states will remain equivalent, in agreement with the index-preservation
property (Eq.~\ref{eq_indexpres}) of Theorem~\ref{theorem1}. The same behavior is observed for a system with two distinct real eigenvalues, whose eigen-separable seeds form a grid in the eigen-coordinates and whose cells are parallelograms (Fig.~\ref{fig1}B and E), and for a pure rotation whose eigenvalues lie on the unit circle, under which the partition is transported rigidly (Fig.~\ref{fig1}C and F), illustrating that index preservation is a property of the construction itself rather than of a particular dynamics; in particular, it requires neither stability nor contraction. This observation illustrates the main result that any two states within the same cell of an abstracted model $T$ are bisimilar to each other.

\paragraph*{The finite abstracted model with bounded approximate bisimulation gap}
We next validate the result about approximate bisimulation stated in 
Theorem~\ref{theorem3} on a
three-dimensional example that activates both the refinement along several
eigen-directions and the truncation of an original abstracted model $T$ into a finite model
$S$. As an illustration, consider Eq.~\ref{eq01} with
\begin{equation*}
   A=\begin{bmatrix}
   0.6000 & -0.3697 & 0.0242\\
   0 & 0.8839 & -1.1490\\
   0 & 0.3536 & -0.1768
   \end{bmatrix},\qquad
   B=I_3 ,
\end{equation*}
whose spectrum has one real eigenvalue and one complex-conjugate pair. The system is non-normal ($\|A\|=1.52>1$ while the spectral radius
is $\rho(A)=0.60$). Thus, any accuracy estimate expressed through $\|A\|$ would diverge,
whereas the eigenvalue-based power bound (Eq.~\ref{eq_powerbound}) stays valid; the
remaining parameters are listed in the caption of Fig.~\ref{fig:approx}.

Index preservation (Eq.~\ref{eq_indexpres}), on which every exactness result rests,
held in all $5\times10^{5}$ checks ($2\times10^{4}$ random trajectories evolving over
$25$ steps under random inputs, with the cell assignment recomputed from the state
alone at each step). Both
error sources tracked their analytical bounds across the entire sweep
(Fig.~\ref{fig:approx}A and B): the partition error stayed within a constant factor
($0.136$--$0.142$) of the Lemma~\ref{lemma_celldiam} bound as the abstraction grew from
$243$ cells to $3.6\times10^{6}$ cells, and the truncation error decayed at the predicted
geometric rate $\rho(A)$, remaining below the bound $\tau$ by factors between $6$ and $11$ for the $K_1$ sweep and between
$2$ and $4$ for the $K_2$ sweep. The upper bounds are clearly conservative but remain valid, and the slack arose from the worst-case
constants $M$ and $\underline{\alpha}$ that random trajectories do not attain. Sweeping
the partition and horizon parameters traced a Pareto front spanning nearly nine orders of
magnitude in the model size and more than one order of magnitude in accuracy (Fig.~\ref{fig:approx}C). Finally, increasing the
eigenvector conditioning ($M$ from $2.0$ to $20.0$ at fixed $\rho(A)$) inflated the
certificate $\tau'$ from $1.7$ to $29$ while the measured error grew only from $0.48$
to $0.59$ (Fig.~\ref{fig:approx}D), confirming that the eigenvector conditioning $M$, which we report alongside the
precision together with $\operatorname{cond}(\Pi)$, governs the conservatism of the certificate rather than the behavior of
the system.

\subsection*{Mapping different operational scenarios to relevant specifications}
An abstracted model $T$ is built solely from a given discrete-time dynamics. To apply this model to a specific operational scenario, it is necessary to introduce relevant operational specifications over individual cells of $T$, captured by a labeling map $\mathcal{L}$ that assigns to each cell, for every atomic specification, the Boolean value 1 if the specification holds on that cell and 0 otherwise (Materials and Methods).
Changing the scenario therefore changes the map $\mathcal{L}$ only, not the abstracted model. Each specification describes either a constraint or a requirement on the accessibility of the relevant part of the workspace, and a {\em scenario} is a collection of specifications constructed in the following three steps.

First, every obstacle present in the scenario is described by a collection of cells with sufficient margins to absorb
the bounded end-effector tracking error of the vendor servo loop. Any forbidden region of the workspace is captured by the collection of cells whose arm configurations reach into that region, determined from the forward kinematics. All such cells are labeled as being \textsf{unsafe}. Others are labeled as being \textsf{safe}. Second, the mission
target, i.e.,  the grasp configuration at which the finger primitive is to be invoked, is
labeled \textsf{goal} only for cells \textit{contained} entirely in the target region. Third, a mission objective is written as a (linear) temporal-logic formula
over those labeled cells. As an illustration, a single-target mission scenario exercised in our
experiments is captured by the following reach--avoid formula
\begin{equation*}
   \phi \;=\; \big(\mathbf{G}\,\neg\,\textsf{unsafe}\big)\;\wedge\;\big(\mathbf{F}\,\textsf{goal}\big),
\end{equation*}
denoting ``never enter an unsafe configuration, and eventually reach a grasp
configuration''. 

The three case studies reported below indicate that defining each specification over cells of the abstracted model $T$, with temporal information captured by temporal logic formulas, is sufficiently general to describe commonly encountered object grasping tasks. What differs between two scenarios is confined to the definitions of $\mathcal{L}$ and $\phi$, which can be quickly updated in practice. The reuse of the model $T$ saves a user from costly model adaptation: the model is built only once, and a new scenario is re-encoded on it by relabeling alone, whereas a task-coupled abstracted model would have to be rebuilt for every new scenario at a cost comparable to that one-time construction.

\subsection*{A multi-purpose mission planning and execution framework}
\begin{figure}
	\centering
	\includegraphics[width=\textwidth]{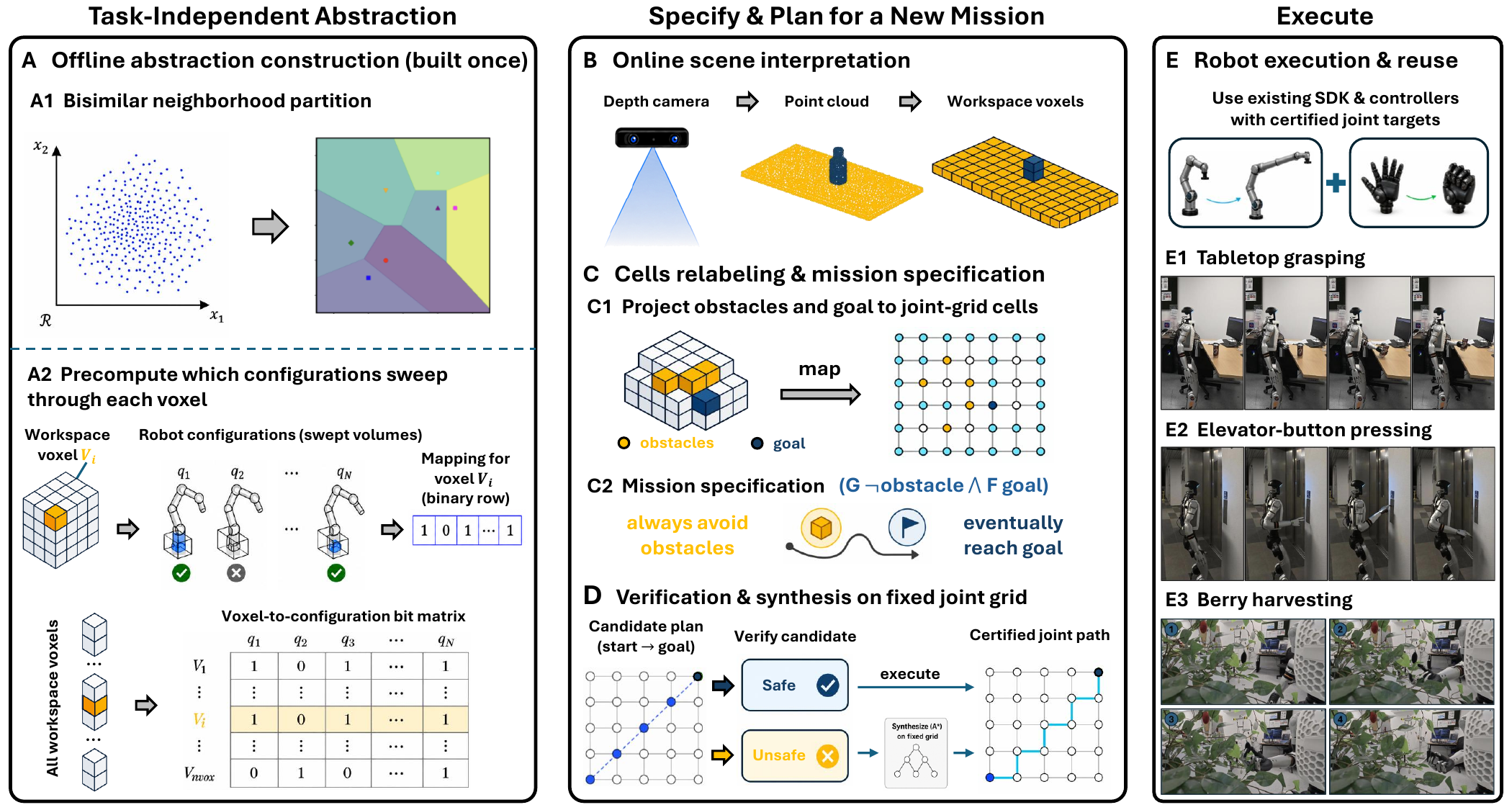}
	\caption{\textbf{The mission planning and execution pipeline, from one-time
		abstraction to certified execution.}
		(\textbf{A})~Offline and task-independent, the abstraction is constructed once
		per robot: the bisimilar neighborhood partition of the state space
		(\textbf{A1}) and a precomputed voxel-to-configuration bit matrix recording
		which arm configurations sweep through each workspace voxel (\textbf{A2}).
		(\textbf{B})~Online, a depth camera converts the current scene into a point
		cloud and then into workspace voxels.
		(\textbf{C})~The occupied and target voxels are projected onto the fixed
		joint-grid cells (\textbf{C1}), and the mission is stated as a temporal-logic
		specification such as
		$\mathbf{G}\,\neg\textsf{unsafe}\wedge\mathbf{F}\,\textsf{goal}$
		(\textbf{C2}).
		(\textbf{D})~A candidate plan is verified cell by cell on the fixed joint grid
		before any motion is commanded; if it is unsafe, or no candidate is offered, a
		certified joint path is synthesized by $A^{\ast}$ search.
		(\textbf{E})~The certified joint targets are executed through the vendor's
		existing SDK and controllers, and the same abstraction is reused across the
		three hardware missions: tabletop grasping (\textbf{E1}), elevator-button
		pressing (\textbf{E2}), and berry harvesting (\textbf{E3}).}
	\label{fig:pipeline}
\end{figure}

\begin{figure}
	\centering
	\includegraphics[width=\textwidth]{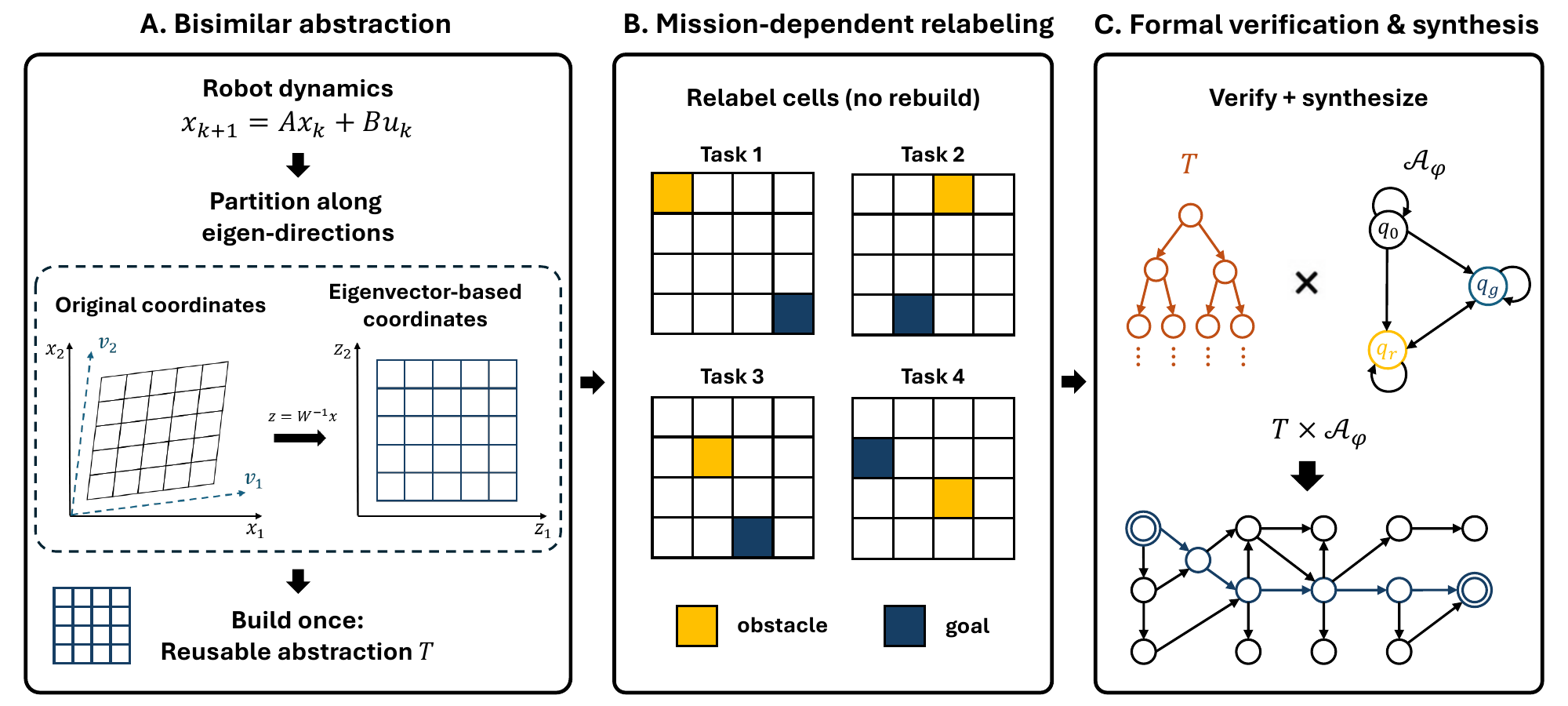}
	\caption{\textbf{One analytically built model, reused across missions.}
		(\textbf{A})~Bisimilar abstraction. The abstraction is built once from the robot dynamics
		$x_{k+1}=Ax_k+Bu_k$: partitioning the state space along the eigen-directions of
		the system matrix becomes, in the eigenvector-based coordinates $z=W^{-1}x$, a
		uniform grid, yielding the reusable abstraction $T$.
		(\textbf{B})~Mission-dependent relabeling. A new mission is encoded by relabeling the existing cells with
		its own obstacle (yellow) and goal (blue) regions, never by rebuilding the model;
		four different tasks share the same partition.
		(\textbf{C})~Formal verification and synthesis. The relabeled model is composed with the specification
		automaton $\mathcal{A}_{\varphi}$, and verification and synthesis on the product
		$T\times\mathcal{A}_{\varphi}$ return a certified plan (blue path), which is
		executed through the platform's existing arm-stretching and finger primitives.
		On the Unitree G1 U1, one such model serves all three missions reported here.}
	\label{fig:concept}
\end{figure}

We combine an abstracted model $T$ built once and real-time adjustable specifications $(\mathcal{L},\phi)$ into a single mission-planning pipeline (Fig.~\ref{fig:pipeline}). A candidate mission plan, for instance, a joint-space path that the platform's own
reaching primitives would follow to the grasp pose, is first \textit{verified} on the
abstracted model $T$. That is, the plan is mapped to the sequence of cells that it visits and checked against
$\mathcal{L}$ and $\phi$. If it visits an \textsf{unsafe} cell the plan is rejected \textit{before} any
motion is commanded, which is the failure mode that the vendor primitives cannot
detect on their own, since each individual reach is executed correctly while their
composition collides. In case the given mission plan candidate is rejected or no candidate is offered, a formal plan synthesis procedure will be carried out, which, by searching the product of the model labeled by $\mathcal{L}$ with the specification automaton
of $\phi$, using for example an $A^{\ast}$ search algorithm on the grid with
an admissible heuristic, returns a certified input sequence.

The synthesized sequence is then mapped back to the physical arm and \textit{executed by the platform's existing functions}. More explicitly, each symbolic transition in $T$ becomes one arm-stretching command to the corresponding joint configuration, and
the terminal transition into the \textsf{goal} cell is followed by a finger-closing command.
No new controller is introduced at execution time. Soundness on hardware follows from
Theorem~\ref{theorem2} together with the margin argument of Materials and Methods. Because
the abstracted model is finite and enforces an exact bisimulation relation over trajectories in the user-level model (Eq.~\ref{cobot_dynamic}), and
every obstacle is inflated by the bounded tracking error $e_{\max}$, the executed motion is
collision-free provided that the vendor servo loop keeps its end-effector error within $e_{\max}$, which can be verified in practice; the scheme is therefore implementable.

The pipeline is multi-purpose; its effectiveness is illustrated in three experimental scenarios (Fig.~\ref{fig:pipeline}E), where the same robot is applied across different environments and specifications.

\subsection*{Experimental studies}

\subsubsection*{Platform and system setup}
Experiments were carried out on a Unitree G1 U1 humanoid robot, which offers a large
configuration space and mobile-manipulation capability in comparatively unstructured
environments. The arm is commanded through the joint-position interface of the vendor's Software
Development Kit (SDK), and the same SDK supplies the two primitives that the planner invokes, namely arm stretching
to a commanded joint configuration and finger opening and closing. Object and obstacle
poses, together with the candidate grasp pose proposed by a learned grasp-pose network, are
supplied by the platform's perception stack, which lies outside the scope of
the contribution. Our proposed mission planner consumes the resulting occupancy description and target pose
as inputs.

At the user level, the platform is described by the integrator model
(Eq.~\ref{cobot_dynamic}) with $A=I$ and $B=I$, so, as established in Materials and Methods,
the eigen-direction partition of Definition~\ref{definition2} reduces exactly to a uniform
joint-space grid with $\Pi=I$, $\underline{\alpha}=1$ and $M=1$, the cell size being bounded
by Lemma~\ref{lemma_celldiam}. The workspace is voxelized at $1$~cm; the end-effector
tracking error of the vendor servo loop was measured as $e_{\max}=1$~cm, and every obstacle
voxel is inflated by one layer accordingly, which is required for the certificate to transfer to
hardware. Since $\rho(A)=1$, all three missions use the bisimilar abstraction $T$ over
a finite horizon. The joint-space grid and its per-voxel collision map are built once,
offline, in $50$~min; thereafter each mission is encoded by relabeling alone, at
about $10$~ms for the reachable-set recomputation and on the order of $100$~ms for the
$A^{\ast}$ synthesis.

Three missions with increasing difficulty levels are reported: object grasping, elevator-button
pressing, and berry harvesting, spanning from tabletop manipulation, interaction with built
infrastructure to  manipulation inside dense, irregular foliage. All three missions are executed on the
same robot with the same abstracted model, joint-space quantization, and planning pipeline.
They differ only in the specification and in the labeling of obstacle and target regions. Figure~\ref{fig:g1_results} shows the three runs, and Movie~S1 the
corresponding footage.

\subsubsection*{Case 1: object grasping}
The first mission requires the robot to reach an object on a tabletop and close its fingers
on it while keeping the whole arm clear of the surrounding clutter. Cells whose inflated
workspace volume intersects the clutter are labeled \textsf{unsafe} and the cells contained
in the grasp region are labeled \textsf{goal}, giving the reach--avoid formula $\phi$ of the
preceding section. The candidate reach produced by the platform's own stretching function
was verified against $\phi$ on the relabeled model prior to execution; a
certified control sequence was then synthesized and executed as a sequence of arm-stretching
commands terminated by a finger-closing command. In the reported trial, the robot completed the mission
collision-free (Fig.~\ref{fig:g1_results}A). In this baseline case, the guarantee is
obtained without modifying the vendor primitives, and the cost of each mission consists of 
relabeling and search only, which are negligible.

\subsubsection*{Case 2: elevator-button pressing}
The second mission requires the robot to press an elevator call button, a task in which
the target is a small planar region on a wall and the surrounding
structure leaves little clearance for the forearm and elbow. Only the labeling changed
compared to Case~1: the button face defines the \textsf{goal} cells, and the wall, the door frame
and the area surrounding the panel define the \textsf{unsafe} cells. Neither the model nor the verification
and synthesis procedures were changed, and no part of the grid was rebuilt. Candidate
motions were again verified before execution and a certified sequence synthesized, which the
robot executed successfully in the reported trial, pressing the button while satisfying the clearance
specification (Fig.~\ref{fig:g1_results}B). This case shows most clearly that each
individual reach is executed correctly by the vendor stack, yet the whole-arm trajectory
has to be certified for the mission to be safe.

\subsubsection*{Case 3: berry harvesting}
The third mission requires the robot to harvest a berry among foliage and supporting
structure, a scene whose obstacles are thin, irregular and densely interleaved with the
target rather than flat and man-made as in the two preceding cases. As before, only the specification and the labels changed:
the berry defines the \textsf{goal} region and the surrounding branches and trellis the
\textsf{unsafe} region, with the same one-voxel inflation absorbing the tracking error.
Verification of the candidate motion and synthesis of a certified control sequence proceeded
exactly as in the two preceding cases, and in the reported trial the robot completed the harvest while avoiding the
obstacles and satisfying its specification (Fig.~\ref{fig:g1_results}C).

In summary, the three missions, in a tabletop scene, a built environment
and a cluttered plant canopy, were certified and executed from one model, built once
from the robot's own dynamics and never rebuilt afterwards. In every case, the per-mission cost
consisted of relabeling and search, about four orders of magnitude below the $50$-min construction
of the model, which was carried out once, whereas a task-coupled abstraction would have had to be
rebuilt for each of the three missions.

\section*{Discussion}
The distinctive feature of this new robotic mission planning framework is that there exists one abstracted model bisimilar to the original dynamics, which is built only once but can be reused for a variety of different application scenarios, avoiding costly reconstruction of the model when the operational environment changes. Aligning the cells with the flow of the dynamics makes
the abstracted model \textit{deterministic}, whereas grid-based symbolic models that re-quantize the state at every
step typically yield nondeterministic and only approximately bisimilar transitions~\cite{%
girard2007approximation,reissig2017feedback}. The idea of reusing one abstracted model across
specifications is also adopted in existing feedback-refinement
methods~\cite{reissig2017feedback,rungger2016scots}. The latter, however, usually result in nondeterministic abstracted models, incurring significantly higher construction complexity. In addition, because the abstract states of such models are related to the original system only by an alternating simulation or an approximate bisimulation rather than an exact one, their guarantees transfer to the original system at the price of a conservatism that grows with the quantization error, which causes more conservative compensations in actual deployment. Finally, by separating different application scenarios from an abstracted model built once and modeling them formally, each resulting mission plan on top of existing system functional primitives can be certifiably derived rather than learned from data.

Decoupling the model from the operational tasks reduces the cost of maintaining a
guarantee over a robot's lifetime. A robot that performs many tasks, or a fleet of robots
sharing one certified controller, would otherwise have to be re-certified whenever the
environment or the rules change. In our framework this cost is confined to relabeling
existing cells with new properties. The separation also provides an interface between
learning and formal methods: perception and task specification, which learning handles
well, enter only through the cell labels, while the analytic core retains its guarantees.
The certified part of the system therefore remains small, which simplifies its safety
assessment.

This study has two main limitations. The hardware evidence is confined to three missions
on a single embodiment, each executed once, and since the servo-level model has
$\rho(A)=1$, it exercises only the bisimilar construction with $A=I$; the general
eigen-direction partition and the approximate bisimilar abstraction $S$ are validated in simulation
only. Moreover, the construction requires linear dynamics with a diagonalizable and
nonsingular system matrix (Assumption~\ref{assumption_diag}), and the Koopman extension
to nonlinear dynamics is only sketched and not exercised by the present experiments.

These limitations suggest directions for future work. In terms of modeling
complexity, the finite abstracted model $S$ has about $J(K_1+1)|U|^{K_2}$ states,
allowing a user to determine a good tradeoff between computational complexity and
acceptable model inaccuracy reflected in the approximate bisimulation relation; a
systematic guidance for such a tradeoff is left to future work. Nonsingularity of $A$
is automatic for sampled-data models $A=e^{A_c\Delta t}$, and the excluded defective
case can be approached through the $\epsilon$-Jordan form, which suggests an
approximately invariant partition with an explicit robustness margin at the price of a
basis conditioning that might deteriorate for large Jordan blocks. For the Koopman-based treatment of nonlinear
dynamics, a bound on the lifting residual and a Lipschitz decoder to ensure a
state-space precision are required, which we leave to future work, along with
tightening the bounds for high-dimensional lifted models.

\section*{Materials and Methods}
This section states the model, definitions, and the main results. All proofs, together
with supporting technical remarks, are given in the Supplementary Text.

\subsection*{Transition systems and problem formulation}
Let $\mathbb{R}^{n}$ and $\mathbb{R}^{m\times n}$ denote the sets of $n$-dimensional
real vectors and $m\times n$ real matrices; $\mathbb{C}$ and $\mathbb{C}^{n}$ the
complex numbers and $n$-dimensional complex vectors; $\mathbb{N}$ the natural
numbers; $\mathbf{0}$ an all-zero vector or matrix of compatible dimension; and
$|\cdot|$, $\|\cdot\|$ the modulus of a complex number and the Euclidean norm. For a
complex vector $c$ and a real vector $x$, $c^H$ is the conjugate transpose and
$\langle c,x\rangle=c^Hx$ the inner product. Given a matrix $X$, we use $\operatorname{spec}(X)$ to denote its spectrum and $\rho(X):=\max_{\mu\in\operatorname{spec}(X)}|\mu|$ for its spectral radius.

\begin{definition}A transition system (with observations) is a tuple
$T=(V,V_0, U, \delta,O,h)$, where $V$ is a (possibly infinite) set of states,
$V_0\subseteq V$ a set of initial states, $U$ a set of inputs,
$\delta: V \times U \rightarrow V$ a transition function, $O$ a set of
observations, and $h: V \rightarrow O$ an observation map. It is finite if $V$ is
finite.
\end{definition}
Let $U^*$ denote the free monoid of $U$ with the string concatenation as the binary operation and $\epsilon$ as the unit element of concatenation, representing an empty string such that for all $\xi\in U^*$, $\epsilon \xi = \xi\epsilon = \xi$. For each $k\in\mathbb{N}$, let $U^k:=\{\sigma_1\cdots \sigma_k|(\forall p:1\leq p\leq k)\sigma_p\in U\}$ for $k>0$, and $U^0:=\{\epsilon\}$. For each $\xi\in U^*$, we use $|\xi|$ to denote the number of non-empty symbols in $\xi$. The transition function is extended to input sequences $\xi\in U^*$ in the usual way, $\delta(v,\epsilon)=v$ and $\delta(v,\xi u)=\delta(\delta(v,\xi),u)$, and we write $\delta(v,\xi)!$ to mean that $\delta(v,\xi)$ is defined. In this paper, $\delta$ is assumed total on $U$, i.e., $\delta(v,\xi)!$ holds for every $\xi$. The notation is kept for uniformity with general systems, whose transition functions may be partial.

Consider a discrete-time linear system
\begin{equation}
\label{eq01}
      x_{k+1}=Ax_k+Bu_k,
\end{equation}
where $x_k\in\mathbb{R}^{n}$, $A\in\mathbb{R}^{n\times n}$,
$B\in\mathbb{R}^{n\times m}$. The set of initial states $\mathcal{I}$ is a compact
subset of $\mathbb{R}^n$ with $x_0\in \mathcal{I}$, and the input set $U$ is a finite
subset of $\mathbb{R}^m$ containing $\mathbf{0}$, denoted $U=\{u_0,u_1,\dots,u_L\}$
with $u_0=\mathbf{0}$. The system (Eq.~\ref{eq01}) is the transition system
$\Sigma=(\mathbb{R}^n, \mathcal{I}, U,\delta,\mathbb{R}^n,h)$, where
$\delta(x,u)=x^{\prime}$ if  $x^{\prime}=Ax+Bu$, and the observation map is
$h(x)=x$. We construct an abstraction $T$ for (Eq.~\ref{eq01}) by discretizing
the state space into a set of quotient states, independent of any proposition,
and define a projection from the continuous state space to the quotient space
so that the abstraction is bisimilar to (Eq.~\ref{eq01}), realizing the commutative
diagram of fig.~\ref{projection}.

\subsection*{Eigen-directions and the partitioning metric}
We partition the state space along the eigen-directions of $A$. Each single
eigen-direction induces only a seminorm and is metrically degenerate when the dimension of the space is
two and above. Therefore, we propose to use a set of eigen-directions, of which the single-direction
case is the coarsest member.

\begin{assumption}
   \label{assumption_diag}
   The matrix $A$ is diagonalizable over $\mathbb{C}$ and nonsingular, i.e.,
   $0\notin\operatorname{spec}(A)$.
\end{assumption}

Under Assumption~\ref{assumption_diag}, taking a basis of left eigenvectors of $A$ and
keeping exactly one member of each complex-conjugate pair yields a set of
linearly independent \textit{eigen-directions} $C=\{c_1,\ldots,c_r\}$ with $c_i^H A=\mu_i c_i^H$. If $A$
has $n_{\mathrm{r}}$ real eigenvalues and $n_{\mathrm{c}}$ complex-conjugate pairs,
then $r=n_{\mathrm{r}}+n_{\mathrm{c}}$, so $\lceil n/2\rceil\leq r\leq n$. Writing
$c_i^H=a_i^T+\mathrm{i}b_i^T$ with $a_i,b_i\in\mathbb{R}^n$, and
$\mu_i=\gamma_i+\mathrm{i}\omega_i$, one has for every real vector $z$
\begin{equation}
   \label{eq_seminorm}
   |\langle c_i,z\rangle|=\sqrt{(a_i^Tz)^2+(b_i^Tz)^2}.
\end{equation}
Define the \textit{eigen-coordinate} mappings
\begin{equation}
   \label{eq_eigencoord}
   \pi_i(z)=\begin{cases} a_i^Tz\in\mathbb{R}, & \mu_i\in\mathbb{R},\\[2pt]
   \big[a_i^Tz,\ b_i^Tz\big]^T\in\mathbb{R}^{2}, & \mu_i\notin\mathbb{R},\end{cases}
\end{equation}
so that $|\langle c_i,z\rangle|=\|\pi_i(z)\|$, and stack them into
$\Pi z=\big[\pi_1(z)^T,\ldots,\pi_r(z)^T\big]^T$. Since $\pi_i$ has dimension $d_i=1$
when $\mu_i\in\mathbb{R}$ and $d_i=2$ otherwise, $\sum_{i=1}^{r}d_i=n$ and
$\Pi\in\mathbb{R}^{n\times n}$. The \textit{projection distance} along the $i$-th
eigen-direction is
\begin{equation}
   \label{eq_di}
   \dc{i}{x}{y}:=|\langle c_i,x-y\rangle|=\|\pi_i(x)-\pi_i(y)\|,\quad x,y\in\mathbb{R}^n.
\end{equation}
Each $\dc{i}{\cdot}{\cdot}$ is a seminorm, but their aggregate is a metric, as shown in
the following lemma.

\begin{lemma}
   \label{lemma_modal}
   Under Assumption~\ref{assumption_diag}, $\Pi:\mathbb{R}^n\to\mathbb{R}^n$ is a
   linear bijection, and for every $z\in\mathbb{R}^n$ and every $i\in\{1,\ldots,r\}$,
   \begin{equation}
      \label{eq_blockscaling}
      \pi_i(Az)=\Lambda_i\pi_i(z),\qquad \|\Lambda_i\zeta\|=|\mu_i|\,\|\zeta\|~~\forall\zeta\in\mathbb{R}^{d_i},
   \end{equation}
   where $\Lambda_i=\mu_i$ if $\mu_i\in\mathbb{R}$, and
   $\Lambda_i=\left[\begin{smallmatrix}\gamma_i&-\omega_i\\ \omega_i&\gamma_i\end{smallmatrix}\right]$
   otherwise. Moreover, with $\underline{\alpha}:=\sigma_{\min}(\Pi)>0$,
   \begin{equation}
      \label{eq_normequiv}
      \|z\|\leq\frac{1}{\underline{\alpha}}\Big(\sum_{i=1}^{r}|\langle c_i,z\rangle|^2\Big)^{\frac{1}{2}}\leq\frac{\sqrt{r}}{\underline{\alpha}}\max_{1\leq i\leq r}|\langle c_i,z\rangle|.
   \end{equation}
\end{lemma}

Thus, $\max_{1\leq i\leq r}\dc{i}{\cdot}{\cdot}$ is a norm equivalent to the Euclidean
one, which converts any bound on the per-direction seminorms into a Euclidean bound
with constant $1/\underline{\alpha}$. To realize the $r$ one-directional partitions
with a single representative point, the representative set must be compatible with the
eigen-coordinates.

\begin{definition}
\label{definition2}
A finite set $\mathbb{V}\subseteq\mathbb{R}^n$ is \textit{eigen-separable} if
$\Pi(\mathbb{V})=G_1\times\cdots\times G_r$ for finite sets
$G_i\subseteq\mathbb{R}^{d_i}$. Given such a $\mathbb{V}$, the neighborhood cell of
each $v\in\mathbb{V}$ is
\begin{equation*}
   \begin{aligned}
      N(v)=\big\{x\in\mathbb{R}^n \,\big|\, &\dc{i}{x}{v}\leq\dc{i}{x}{v'}, \forall v'\in\mathbb{V},~\forall i\in\{1,\ldots,r\}\big\},
   \end{aligned}
\end{equation*}
where ties are broken coordinatewise: each finite set $G_i$ is given a fixed total order, and
if several points of $G_i$ are equidistant from $\pi_i(x)$, the smallest one in that order is
selected; $v$ is then the unique element of $\mathbb{V}$ whose eigen-coordinates are the
selected points.
\end{definition}

\begin{lemma}
\label{lemma1}
Consider a compact set of initial states $\mathcal{I}\subseteq\mathbb{R}^n$. For any
$\varepsilon>0$ there exists a finite eigen-separable set
$\mathcal{I}_A\subset \mathbb{R}^n$ such that
\begin{equation}
   \label{eq_cover_multi}
   (\forall x \in \mathcal{I})(\exists v_0 \in \mathcal{I}_A)~\max_{1\leq i\leq r}\dc{i}{x}{v_0} \leq \varepsilon,
\end{equation}
and consequently $\|x-v_0\|\leq\sqrt{r}\,\varepsilon/\underline{\alpha}$.
\end{lemma}

\subsection*{The abstracted transition system}
Based on $\mathcal{I}_A$ from Lemma~\ref{lemma1}, let $J:=|\mathcal{I}_A|$, index the seeds as $\mathcal{I}_A=\{v_0^1,\ldots,v_0^J\}$, write $v_k^j:=A^kv_0^j$, and let $A^k \mathcal{I}_A:=\{A^kv_0|v_0\in\mathcal{I}_A\}=\{v_k^1,\ldots,v_k^J\}$. We define a transition system
$T=(V,V_0, U, \hat{\delta},\mathbb{R}^n,\hat{h})$, where the set of (symbolic) states is
\begin{equation}
   \label{eq02}
   \begin{aligned}
      V=\Big\{\big(v_k,\xi\big)\Big | k \in \mathbb{N},\, v_k\in A^k \mathcal{I}_A,\, k\geq 1\Rightarrow \xi=u_{l_0}\cdots u_{l_{k-1}}\in U^k, k=0\Rightarrow \xi = \epsilon\Big\}.
   \end{aligned}
\end{equation}
The initial symbolic states are
$V_0=\big\{(v_0,\epsilon) \mid v_0\in \mathcal{I}_A\big\}\subset V$ and the transition relation $\hat{\delta}$ is
\begin{equation}
   \label{eq03}
   \begin{aligned}
     (\forall k\in\mathbb{N}) (\forall \xi\in U^k)(\forall (v_k,\xi)\in V) (\forall u\in U) ~&\hat{\delta}\left(\big(v_k,\xi \big),u\right)=\big(Av_k,\xi u\big).
   \end{aligned}
\end{equation}
For all $k\in\mathbb{N}$ and $(v_k,u_{l_0}\cdots u_{l_{k-1}})\in V$, if $k\geq 1$, then
\begin{equation}
   \label{eq04}
   \hat{h}\big(v_k,u_{l_0}\cdots u_{l_{k-1}}\big) =v_k+\sum_{i=0}^{k-1}A^i B u_{l_{k-1-i}},
\end{equation}
and $\hat{h}(v_0,\epsilon)=v_0$ on the initial layer. Equivalently, $\hat{h}$ obeys the one-step recursion
\begin{equation}
   \label{eq_hrec}
   \hat{h}\big(\hat{\delta}(v,u)\big)=A\,\hat{h}(v)+Bu,\qquad \forall v\in V,~\forall u\in U.
\end{equation}
The system $T$ is a
multi-layer tree-structured abstraction (fig.~\ref{tree}), where each directed edge denotes a transition
for a specific input.

\begin{definition}
\label{definition3}
For each $\xi\in U^*$, a projection $P_{\xi}:\mathbb{R}^n \to V$ is defined by $P_{\xi}(x):=v$,
where
\begin{equation}
   \label{eq_proj}
   v\in\bigcap_{i=1}^{r}\ \operatorname*{arg\,min}_{v^{\prime}\in \tilde{V}}\ \dc{i}{x}{\hat{h}(v^{\prime})},
\end{equation}
with $\tilde{V}=\{(v_{k}, \xi) \mid v_k\in A^{k}\mathcal{I}_A\}$, $k=|\xi|$, the layer reached by
the input string $\xi$. Ties are broken in each
eigen-coordinate by the coordinatewise rule of Definition~\ref{definition2}, which
renders $v$ unique. The kernel of $P_{\xi}$ is an equivalence relation on  $\mathbb{R}^n$, denoted as $\equiv_{\xi}$, where 
for all $x,y\in \mathbb{R}^n$,  $x\equiv_{\xi} y \iff P_{\xi}(x) = P_{\xi}(y)$.
\end{definition}

\begin{proposition}
Given  Assumption~\ref{assumption_diag}, the projection $P_{\xi}$ is well defined.
\end{proposition}

\begin{definition}
   \label{definition4}
   Given the transition system $\Sigma=(\mathbb{R}^n, \mathcal{I}, U,\delta,\mathbb{R}^n,h)$ associated with (Eq.~\ref{eq01}) and its abstracted model $T=(V,V_0, U, \hat{\delta},\mathbb{R}^n,\hat{h})$, two states $x,y\in\mathbb{R}^n$ are \textit{bisimilar} with respect to $\xi\in U^{*}$ if (1) $x\equiv_{\xi} y$; (2) for all
   finite input sequences $\eta\in U^{*}$, $\delta(x,\eta)!$ if and only if $\delta(y,\eta)!$,
   and $\delta(x,\eta)\equiv_{\xi\eta} \delta(y,\eta)$.
\end{definition}

That is, if two states are bisimilar, then any trajectories originating from these two states under the same input sequence are confined to the same sequence of cells.

\subsection*{Bisimulation and property certifiability}

\begin{theorem}
\label{theorem1}
Given $\Sigma=(\mathbb{R}^n, \mathcal{I}, U,\delta,\mathbb{R}^n,h)$ and its abstracted model
$T=(V,V_0, U, \hat{\delta},\mathbb{R}^n,\hat{h})$ built, under Assumption~\ref{assumption_diag}, from the
eigen-separable set $\mathcal{I}_A$ of Lemma~\ref{lemma1}, for any states $x,y\in\mathbb{R}^n$
and any $\xi\in U^{*}$, if $x\equiv_{\xi} y$, then $x$ and $y$ are bisimilar with
respect to $\xi$.
\end{theorem}

The core of the proof (Supplementary Text) is the one-step \textit{index-preservation}
property. Write $x^{+}=Ax+Bu$ for the successor of $x$ under (Eq.~\ref{eq01}); one can show that
\begin{equation}
   \label{eq_indexpres}
   P_{\xi}(x)=(v_k^F,\xi)\ \Longrightarrow\ P_{\xi u}(x^{+})=(v_{k+1}^F,\xi u).
\end{equation}
Therefore, the representative index $F$ is carried along the dynamics, and any two states in
the same cell, i.e., being equivalent under $\equiv_{\xi}$,  remain so under the same input string.
Iterating (Eq.~\ref{eq_indexpres}) along an arbitrary input sequence shows that the
projection commutes with the two transition maps, which is the commutative diagram of
fig.~\ref{projection}.

\begin{corollary}
\label{corollary1}
Given $\Sigma=(\mathbb{R}^n, \mathcal{I}, U,\delta,\mathbb{R}^n,h)$ and its abstracted model
$T=(V,V_0, U, \hat{\delta},\mathbb{R}^n,\hat{h})$,
\[(\forall x\in\mathbb{R}^n)(\forall \xi\in U^*)~v=P_{\xi}(x)\Rightarrow(\forall \eta\in U^*)\,\big[\delta(x,\eta)!\Rightarrow P_{\xi\eta}\big(\delta(x,\eta)\big)=\hat{\delta}(v,\eta)\big].\]
\end{corollary}

Conversely, every transition of $T$ is realized by $\Sigma$: the representative point
$v_0\in \mathcal{I}_A$ lies in its own cell $P_{\epsilon}^{-1}((v_0,\epsilon))$ and, by
(Eq.~\ref{eq_hrec}), its trajectory under any $\xi$ is exactly the propagated
representative $\hat{h}(\hat{\delta}((v_0,\epsilon),\xi))$. Hence $\Sigma$ and $T$ are
bisimilar to each other in the sense that any initial state $x\in\mathbb{R}^n$ of $\Sigma$ and the representative point $\hat{h}(P_{\epsilon}(x))$ of its initial cell are bisimilar with respect to $\epsilon$ (Theorem~\ref{theorem1}). Next, we will show that a property verified on $T$ transfers to $\Sigma$ provided it
is decided cellwise, which is formalized below.

A \textit{property} is a map $\Psi:\mathbb{R}^n\times U^*\rightarrow\{0,1\}$, where
$\Psi(x,\xi)=1$ means that the trajectory of (Eq.~\ref{eq01}) starting from $x$ under the
input sequence $\xi$ meets a given requirement, e.g., it reaches a target region while
avoiding obstacles. Since $T$ only distinguishes states up to the cells of the initial
partition, a property can be certified on $T$ only if it is decided cellwise.

\begin{definition}
\label{definition_blockcons}
Given $\Sigma=(\mathbb{R}^n, \mathcal{I}, U,\delta,\mathbb{R}^n,h)$ and its abstracted model
$T=(V,V_0, U, \hat{\delta},\mathbb{R}^n,\hat{h})$, a property $\Psi:\mathbb{R}^n\times U^*\rightarrow\{0,1\}$
is \textit{block conservative} with respect to $T$, if for all $x\in\mathbb{R}^n$ and $\xi\in U^*$,
\[\Psi(x,\xi)=1\Rightarrow(\forall y\in\mathbb{R}^n)\big[x\equiv_{\epsilon} y\Rightarrow\Psi(y,\xi)=1\big].\]
\end{definition}

\begin{definition}
\label{definition_hull}
Given $\Sigma=(\mathbb{R}^n, \mathcal{I}, U,\delta,\mathbb{R}^n,h)$ and its abstracted model
$T=(V,V_0, U, \hat{\delta},\mathbb{R}^n,\hat{h})$, for any property
$\Psi:\mathbb{R}^n\times U^*\rightarrow\{0,1\}$, a block conservative property
$\Psi^{\prime}:\mathbb{R}^n\times U^*\rightarrow\{0,1\}$ is a \textit{conservative hull} of
$\Psi$ with respect to $T$, if for all $x\in\mathbb{R}^n$ and $\xi\in U^*$,
$\Psi^{\prime}(x,\xi)=1\Rightarrow\Psi(x,\xi)=1$. $\Psi^{\prime}$ is \textit{minimum} if for any
other conservative hull $\Psi^{\prime\prime}$ of $\Psi$, we have
$\Psi^{\prime\prime}(x,\xi)=1\Rightarrow\Psi^{\prime}(x,\xi)=1$.
\end{definition}

\begin{theorem}
\label{theorem2}
Given $\Sigma=(\mathbb{R}^n, \mathcal{I}, U,\delta,\mathbb{R}^n,h)$ and its abstracted model
$T=(V,V_0, U, \hat{\delta},\mathbb{R}^n,\hat{h})$, for any property
$\Psi:\mathbb{R}^n\times U^*\rightarrow\{0,1\}$, there exists a unique minimum conservative hull
$\Psi^{\prime}$ of $\Psi$ such that
\[(\forall (v_0,\epsilon)\in V_0)(\forall \xi\in U^*)~\Psi^{\prime}(v_0,\xi)=1\Rightarrow\big(\forall x\in P_{\epsilon}^{-1}((v_0,\epsilon))\big)~\Psi^{\prime}(x,\xi)=1.\]
\end{theorem}

A conservative hull never certifies a pair $(x,\xi)$ that violates $\Psi$, and the minimum
hull is the least conservative one, i.e., it certifies every pair that any cellwise
certificate can certify. Theorem~\ref{theorem2} states that it suffices to check
$\Psi^{\prime}$ at the representative point $v_0\in\mathcal{I}_A$ of a cell: the certificate then extends
to every state of the cell $P_{\epsilon}^{-1}((v_0,\epsilon))$. Since the trajectory of
$v_0$ under $\xi$ is exactly a path of $T$ (Corollary~\ref{corollary1}), a hull that is
defined cellwise, as the labeling rules of the formal verification and control synthesis
subsection construct for reach--avoid specifications, is evaluated on $T$ alone. Hence a
controller for $\Sigma$ that enforces $\Psi$ is obtained directly from the one synthesized on
$T$.

\subsection*{Approximately bisimilar finite abstractions}
Based on the definition of the abstracted model $T$, the finiteness of $T$ is determined by the length of each input sequence. In most applications, only finite horizons are relevant. However, when an infinite planning horizon is under consideration, $T$ contains an infinite number of states. To avoid this theoretical limitation, we propose another model abstraction method based on the concept of {\em approximate bisimulation relation}. To make this method work, we need to impose the following assumption.

\begin{assumption}
\label{assumption_stable}
The linear system (Eq.~\ref{eq01}) is asymptotically stable, that is, $\rho(A)<1$.
\end{assumption}

Under Assumption~\ref{assumption_diag}, the matrix $A$ is diagonalizable, that is, $A=WDW^{-1}$ with $D$ diagonal and $W$ the matrix of right eigenvectors. Then under Assumption~\ref{assumption_stable}, we have
\begin{equation}
   \label{eq_powerbound}
   \big\|A^{k}\big\|\leq M\rho(A)^{k}~~\forall k\in\mathbb{N},\qquad M:=\kappa(W)=\|W\|\big\|W^{-1}\big\|\geq1,
\end{equation}
with $M$ the condition number of the eigenvector basis. Note that $\rho(A)<1$ does not imply
$\|A\|<1$. We introduce two truncation parameters $K_1,K_2\in\mathbb{N}$. For any $k>K_1$, the value of $v_k\in A^k\mathcal{I}_A$ is
replaced by $\mathbf{0}$, and only the most recent $K_2$ controls are retained. Under such approximations, we introduce a
finite transition system $S=(Q,Q_0,U,\tilde{\delta},\mathbb{R}^n,\tilde{h})$, where
\begin{equation}
   \label{eq_Q}
   \begin{aligned}
   Q=\Big\{\big(v_k,u_{l_{k-K_2}}\cdots u_{l_{k-1}}\big)\,\big|~k\in\mathbb{N}, u_{l_{k-K_2}}\cdots u_{l_{k-1}}\in\bigcup_{i=0}^{K_2}U^i \Big\},
   \end{aligned}
\end{equation}
with $v_k\in A^k \mathcal{I}_A$ if $k\leq K_1$ and $v_k=\mathbf{0}$ otherwise, and $u_{l_m}=\epsilon$
if $m<0$. The initial set is $Q_0=\big\{(v_0,\epsilon) \mid v_0\in \mathcal{I}_A\big\}$, and
the transition relation $\tilde{\delta}$ is defined as follows: for all $k\in\mathbb{N}$ and $\big(v_k,u_{l_{k-K_2}}\cdots u_{l_{k-1}}\big)\in Q$,
\begin{equation}
   \label{eq_deltatilde}
   \begin{aligned}
      (\forall u_{l_{k}}\in U)\, \tilde{\delta}\Big(\big(v_k,u_{l_{k-K_2}}\cdots u_{l_{k-1}}\big),u_{l_{k}}\Big) = \big(v_{k+1},u_{l_{k-K_2+1}}\cdots u_{l_{k-1}}u_{l_{k}}\big).
   \end{aligned}
\end{equation}
where $v_{k+1}=Av_k$ if $k+1\leq K_1$ and $v_{k+1}=\mathbf{0}$ otherwise. The observation map is
\begin{equation}
   \label{eq_htilde}
   (\forall k\in\mathbb{N})\, \tilde{h}(v_k,u_{l_{k-K_2}}\cdots u_{l_{k-1}})=v_k+\sum_{i=0}^{K_2-1}A^i B u_{l_{k-1-i}},
\end{equation}
where the terms with $k-1-i<0$ are omitted. Thus, apart from up to $|U|^{K_2}$ states of the merged layers $k>K_1$, which all share the representative $\mathbf{0}$, $S$ has at most $J(K_1+1)|U|^{K_2}$ states regardless of the horizon, while still
admitting the proposition-independent partitioning and projection above. Specifically, the
projection of Definition~\ref{definition3} carries over to $S$ verbatim: for $\xi\in U^*$ of
length $k\in\mathbb{N}$, $\tilde{P}_{\xi}:\mathbb{R}^n\to Q$ selects, within the layer
$\tilde{Q}=\{(v_k,u_{l_{k-K_2}}\cdots u_{l_{k-1}})\mid v_k\in A^k\mathcal{I}_A \text{ if } k\leq K_1,\ v_k=\mathbf{0} \text{ otherwise}\}$ reached by
$\xi$, the state $q$ that minimizes every projection distance $\dc{i}{x}{\tilde{h}(q)}$ with any possible ties
being broken by the coordinatewise rule of Definition~\ref{definition2}.

\begin{definition}
   \label{definition5}
   Given $\Sigma=(\mathbb{R}^n, \mathcal{I}, U,\delta,\mathbb{R}^n,h)$ and its finite abstraction
   $S=(Q,Q_0,U,\tilde{\delta},\mathbb{R}^n,\tilde{h})$, two states $x,y\in\mathbb{R}^n$ are
   \textit{approximately bisimilar} with respect to $\xi\in U^*$ and an approximation error
   $\tau\in\mathbb{R}^{+}$, if (1) $\tilde{P}_{\xi}(x)=\tilde{P}_{\xi}(y)$; (2) for all
   $\eta\in U^*$, $\delta(x,\eta)!$ if and only if $\delta(y,\eta)!$, and either
   $\tilde{P}_{\xi\eta}(\delta(x,\eta))=\tilde{P}_{\xi\eta}(\delta(y,\eta))$ or
   $\|\delta(x,\eta)-\delta(y,\eta)\|\leq\tau$.
\end{definition}

\begin{theorem}
\label{theorem_approx}
Under Assumptions~\ref{assumption_diag} and \ref{assumption_stable}, given
$\Sigma=(\mathbb{R}^n, \mathcal{I}, U,\delta,\mathbb{R}^n,h)$ and its finite abstracted model
$S=(Q,Q_0,U,\tilde{\delta},\mathbb{R}^n,\tilde{h})$, any $x,y\in\mathbb{R}^n$ with
$\tilde{P}_{\xi}(x)=\tilde{P}_{\xi}(y)$ for some $\xi\in U^*$ are approximately bisimilar with
respect to $\xi$ and an approximation error no bigger than
\begin{equation}
   \label{eq_tau}
   \tau=M r_{\mathcal{I}} \rho(A)^{K_1+1}+\frac{M r_U \rho(A)^{K_2}}{1-\rho(A)},
\end{equation}
where $r_{\mathcal{I}}=\max _{v_0\in\mathcal{I}_A}\|v_0\|$ and $r_U=\max _{u \in U}\|Bu\|$.
\end{theorem}

The error $\tau$ is the uniform gap between the observations of $S$ and of the exact
abstraction $T$ along any common input history (Eq.~\ref{eq_seedgap} in the Supplementary
Text); it vanishes as $K_1,K_2\to\infty$, recovering Theorem~\ref{theorem1}.

The following lemma quantifies the gap between an actual state of $\Sigma$ and the corresponding output of $T$, after mapping the actual state to a cell of $T$. 

\begin{lemma}
   \label{lemma_celldiam}
   Under the hypotheses of Theorem~\ref{theorem1} and with covering radius
   $\varepsilon$ as in Lemma~\ref{lemma1}, for any $x_0\in \mathcal{I}$ and input
   string $\xi\in U^{*}$ of length $k$,
   \begin{equation}
      \label{eq_celldiam}
      \big\|x_k-\hat{h}\big(P_{\xi}(x_k)\big)\big\|\leq\big\|A^{k}\big\|\,\frac{\sqrt{r}\,\varepsilon}{\underline{\alpha}}.
   \end{equation}
   Under Assumption~\ref{assumption_stable}, $\|A^k\|\leq M$ for all $k$, so the
   estimate is uniform and equal to $M\sqrt{r}\,\varepsilon/\underline{\alpha}$.
\end{lemma}

\begin{theorem}
\label{theorem3}
Let Assumptions~\ref{assumption_diag} and \ref{assumption_stable} hold and let
$\mathcal{I}_A$ have covering radius $\varepsilon$. For any $x_0\in\mathcal{I}$ with
$P_{\epsilon}(x_0)=(v_0^F,\epsilon)$ and any input string $\xi\in U^*$, the state of $\Sigma$
associated with (Eq.~\ref{eq01}) and the state of $S$ reached from the same initial cell under the same input history satisfy
\begin{equation}
   \label{eq_tauprime_bound}
   \big\|\delta(x_0,\xi)-\tilde{h}\big(\tilde{\delta}((v_0^F,\epsilon),\xi)\big)\big\|\leq\tau^{\prime},
\end{equation}
where
\begin{equation}
   \label{eq_tauprime}
   \tau^{\prime}= \underbrace{\frac{M\sqrt{r}\,\varepsilon}{\underline{\alpha}}}_{\text{abstraction}}+\underbrace{M r_{\mathcal{I}} \rho(A)^{K_1+1}+\frac{M r_U \rho(A)^{K_2}}{1-\rho(A)}}_{\tau,~\text{truncation}}.
\end{equation}
\end{theorem}

The precision (Eq.~\ref{eq_tauprime}) contains two components: the first term is due to abstraction from $\Sigma$ to $T$, and the second term is due to the approximation of $T$ by $S$ via truncation. These two components can be tuned independently, and any target accuracy can be attained \textit{a~priori} by adjusting the granularity of the initial eigen-separable set $\mathcal{I}_A$ and the truncation parameters $(K_1,K_2)$, at the price of incurring $J(K_1+1)|U|^{K_2}$ symbolic states.

\subsection*{Abstraction construction}
Given (Eq.~\ref{eq01}), we extract the eigen-directions $C=\{c_1,\dots,c_r\}$ of $A$
and the eigen-coordinate mapping $\Pi$, discretize $\mathcal{I}$ by a finite
eigen-separable representative set $\mathcal{I}_A$ (following Lemma~\ref{lemma1}), and partition the state space into
neighborhood cells, each corresponding to a symbolic state of $T$. Any continuous state $x$ is mapped
to its representative via $P_{\xi}(x)$. As the system evolves, each representative
point propagates under the dynamics, generating successive layers of symbolic states
indexed by input sequences, which for finite control sequences yields a
tree-structured transition system. The successor of a symbolic state under an input is
fully determined by propagating its representative point. Therefore, $T$ can be generated
straightforwardly along each explored input sequence; peak storage then grows linearly along a
single explored path, though the number of nodes to be searched grows. The size of the truncated model $S$ from $T$,
$J(K_1+1)|U|^{K_2}$, will remain bounded, even for an infinite planning horizon.

\subsection*{Formal verification and control synthesis}
Given (Eq.~\ref{eq01}) and a specification $\phi$ in LTL, the framework proceeds as
follows. (i)~\textit{Abstraction construction:} build $T$ or $S$ by the
neighborhood-based partitioning above; being proposition-independent, the same
abstraction is reused across specifications. (ii)~\textit{Specification encoding:}
attach the propositions \textit{a~posteriori} by a labeling map $\mathcal{L}$. To make the
labeled specification a conservative hull in the sense of Theorem~\ref{theorem2}, safety
(avoid) propositions are \textit{over-%
approximated} (a cell is labeled unsafe if it \textit{intersects} an unsafe region),
while reach (goal) propositions are \textit{under-approximated} (a cell is labeled a
goal only if it is \textit{contained} in the target). Both rules make each
proposition a union of cells, at a conservatism of at most one cell radius, and $\phi$
is evaluated over these labels. (iii)~\textit{Verification:} export $T$ or $S$ with the labeling
into an off-the-shelf model checker (e.g. NuSMV) for automaton-based model checking.
(iv)~\textit{Control synthesis:} construct the product automaton and synthesize an
input sequence satisfying $\phi$, mapped back to the original system through the
projection $P_{\xi}$ (Corollary~\ref{corollary1}). For reach--avoid specifications this is a graph search
(e.g., $A^{\ast}$ with an admissible heuristic) on the product with a deterministic
finite automaton; for general $\omega$-regular specifications an accepting lasso is
sought in the product of $S$ with a generalized B\"uchi automaton. Because the
labeling is decoupled from the abstraction, modifying the specification only requires
relabeling the existing cells; when $S$ is used, the guarantee holds up to the
precision $\tau'$ of Theorem~\ref{theorem3}.

\subsection*{Extension to nonlinear systems}
For a nonlinear discrete-time system in which the state $x_k$ and the control signal $u_k$ enter additively,
\begin{equation}
   \label{eq_nonlinear}
   x_{k+1}=F(x_k)+\beta(u_k),
\end{equation}
we employ the Koopman operator to lift (Eq.~\ref{eq_nonlinear}) into a
higher-dimensional space of observables $\Phi(\cdot)$ in which the dynamics become
linear,
\begin{equation}
   \label{eq_koopman}
   \Phi(x_{k+1})=A_{\Phi}\Phi(x_k)+\psi(u_k),
\end{equation}
where $\Phi$, $A_{\Phi}$ and $\psi$ are identified by data-driven methods, for instance,
deep encoder--decoder networks approximating the Koopman
eigenfunctions~\cite{lusch2018deep}. The lifting is used only in the abstraction
construction: the eigen-directions $C_{\Phi}$ of $A_{\Phi}$ induce on the lifted state
projection distances that contract by the constant factors $|\mu_{\Phi,i}|$, whose
maximum is a genuine metric on the lifted state, and each continuous state $x$ is
projected to $v=P_{\Phi}(x)$ by
\begin{equation}
   \label{eq_koopman_proj}
   \dc{\Phi,i}{\Phi(x)}{\Phi(v)}=\min_{v'\in \mathcal{I}_A}\dc{\Phi,i}{\Phi(x)}{\Phi(v')},\quad \forall i=1,\ldots,r_{\Phi}.
\end{equation}
Apart from this partitioning step, the abstraction is constructed exactly as above, the
representative points being propagated under the actual nonlinear dynamics
(Eq.~\ref{eq_nonlinear}). The guarantees of Theorems~\ref{theorem1}--\ref{theorem3}
carry over only when $A_{\Phi}$ admits an \textit{exact} finite-dimensional invariant
subspace; for a data-driven $A_{\Phi}$ with a nonzero residual the eigen-coordinate
distances contract only approximately, so exactness and index-preservation hold only up
to that residual, and recovering a state-space rather than lifted-space precision
additionally requires a Lipschitz bound on the decoder. A detailed treatment of
this residual is left to future work.

\subsection*{Application to robotic systems}
For a robotic manipulator, the joint-space dynamics are typically nonlinear~\cite{Spong2006Robot},
\[D(q)\ddot{q}+C(q,\dot{q})\dot{q}+\mathbf{g}(q)=\tau_{\mathrm{q}},\]
with inertia matrix $D(q)$, Coriolis matrix $C(q,\dot{q})$, gravity vector $\mathbf{g}(q)$ and joint torque $\tau_{\mathrm{q}}$ (symbols local to this paragraph). But modern
controllers expose only high-level interfaces for querying the joint configuration and
executing joint-space motion commands, with an embedded servo controller performing
torque control, gravity compensation and feedback linearization. This is the interface to which our
mission planner writes: the arm-stretching primitive consumes a commanded joint
configuration and the finger primitive takes a binary open/close command, and neither is modified here.
Let the measured configuration at step $k$ be
$\theta(k)=[\theta_1,\ldots,\theta_{n_\theta}]^T$ and the commanded
increment $u(k)=\theta_{\mathrm{target}}-\theta(k)$. Assuming the servo tracks the
command within one sampling interval, the user-level dynamics are
\begin{equation}
\label{cobot_dynamic}
\theta(k+1)=\theta(k)+u(k),
\end{equation}
a discrete-time linear integrator with $A=I$ and $B=I$. Real position tracking is exact
only up to a bounded end-effector error, measured as $e_{\max}=1$~cm on the G1. Because
obstacles are represented in the workspace as $1$-cm voxels, inflating every obstacle by
one voxel layer ($e_{\max}$) restores soundness: the synthesized motion is guaranteed
collision-free provided the executed end-effector error stays within $e_{\max}$, which
turns the idealized integrator into an approximate bisimulation of workspace precision
$e_{\max}$. When $A=I$, every nonzero vector is a left eigenvector with $\mu=1$; taking the
canonical basis gives $\Pi=I$,
$\underline{\alpha}=1$, $M=1$, and the neighborhoods of Definition~\ref{definition2}
reduce exactly to a uniform grid, which is the joint-space quantization used in the
robot experiments. Since $\rho(A)=1$, the servo-level models are covered by the exactly
bisimilar abstraction $T$ over any finite horizon (which is what reach--avoid
manipulation missions require), instead of the unbounded-horizon abstraction $S$. For the latter, an inner
loop placing the closed-loop poles strictly inside the unit circle is needed to enforce
Assumption~\ref{assumption_stable}. The
same reasoning applies to any platform whose SDK exposes a joint-space or end-effector
position interface, so the model form (Eq.~\ref{cobot_dynamic}) is not specific to the
Unitree G1 U1 used here; only the dimension $n_\theta$ of the configuration space and the
joint limits change. Other nonlinear platforms such as a differential-drive
mobile robot require the Koopman lifting, which is not exercised by the
present experiments.

\begin{figure}
	\centering
	\begin{minipage}[t]{0.32\textwidth}\centering
		\textbf{A}\\[2pt]\includegraphics[width=\linewidth]{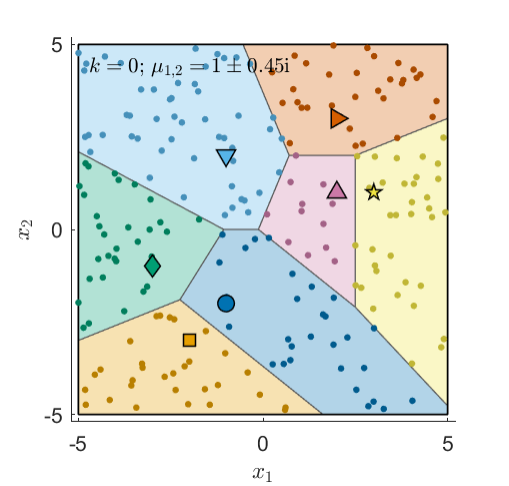}
	\end{minipage}\hfill
	\begin{minipage}[t]{0.32\textwidth}\centering
		\textbf{B}\\[2pt]\includegraphics[width=\linewidth]{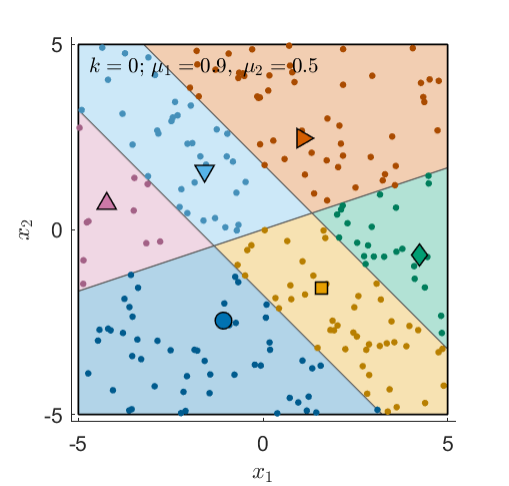}
	\end{minipage}\hfill
	\begin{minipage}[t]{0.32\textwidth}\centering
		\textbf{C}\\[2pt]\includegraphics[width=\linewidth]{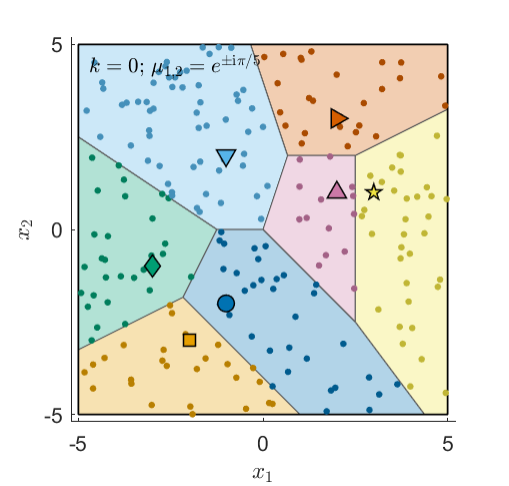}
	\end{minipage}\\[6pt]
	\begin{minipage}[t]{0.32\textwidth}\centering
		\textbf{D}\\[2pt]\includegraphics[width=\linewidth]{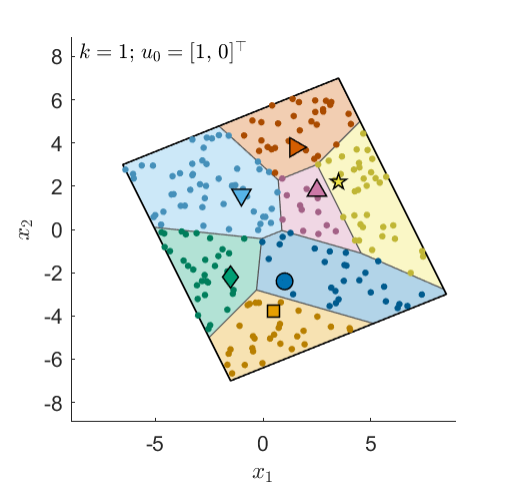}
	\end{minipage}\hfill
	\begin{minipage}[t]{0.32\textwidth}\centering
		\textbf{E}\\[2pt]\includegraphics[width=\linewidth]{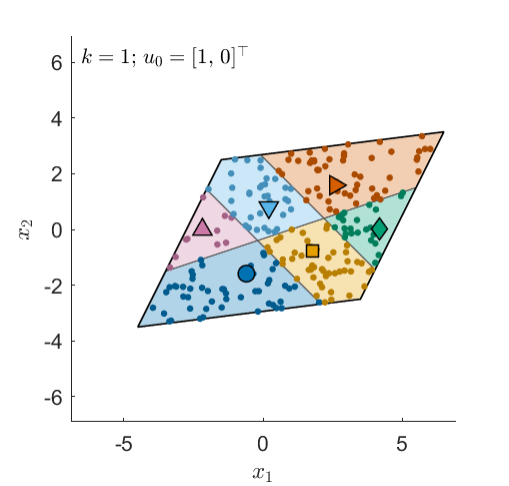}
	\end{minipage}\hfill
	\begin{minipage}[t]{0.32\textwidth}\centering
		\textbf{F}\\[2pt]\includegraphics[width=\linewidth]{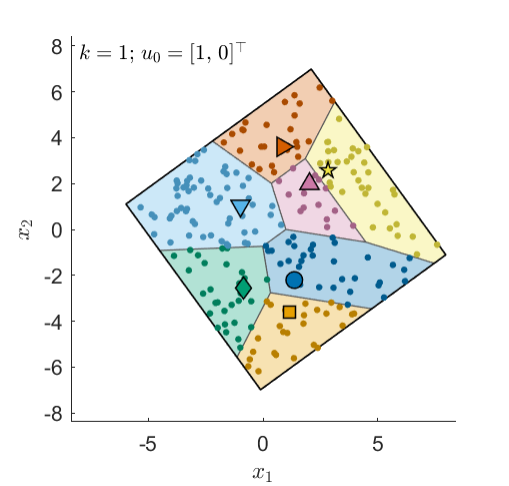}
	\end{minipage}
	\caption{\textbf{Validation of the proposed neighborhood partition across
		different dynamics.}
		Each column shows one discrete-time linear system, all diagonalizable and
		nonsingular:
		(\textbf{A} and \textbf{D})~the worked example
		$A=\left[\begin{smallmatrix}1&-0.5\\0.4&1\end{smallmatrix}\right]$ with the
		complex pair $\mu_{1,2}=1\pm0.45\mathrm{i}$;
		(\textbf{B} and \textbf{E})~$A=\left[\begin{smallmatrix}0.8&0.3\\0.1&0.6\end{smallmatrix}\right]$
		with distinct real eigenvalues $\mu_1=0.9$ and $\mu_2=0.5$, for which the
		eigen-separable seeds form a grid in the eigen-coordinates and the cells are
		parallelograms;
		(\textbf{C} and \textbf{F})~a pure rotation by $\pi/5$,
		$A=\left[\begin{smallmatrix}0.81&-0.59\\0.59&0.81\end{smallmatrix}\right]$,
		whose complex pair $\mu_{1,2}=e^{\pm\mathrm{i}\pi/5}$ lies on the unit circle
		($|\mu_{1,2}|=1$), so the partition is transported rigidly, with neither
		contraction nor expansion.
		The top row shows, at $k=0$, the partition induced by the seeds under the
		eigen-direction metric, together with random states colored by the cell that
		contains them. The bottom row shows the situation after the common input
		$u=[1,\,0]^{T}$: the seeds propagate to $Av+Bu$, inducing the offspring
		partition, and every state, propagated to $Ax+Bu$, keeps its color, that is,
		states originating from the same cell remain within the corresponding
		propagated cell, visualizing the index-preservation property
		(Eq.~\ref{eq_indexpres}) of Theorem~\ref{theorem1} for each of the three
		dynamics.}
	\label{fig1}
\end{figure}

\begin{figure}
	\centering
	\begin{minipage}[t]{0.38\textwidth}\centering
		\textbf{A}\\[2pt]\includegraphics[width=\linewidth]{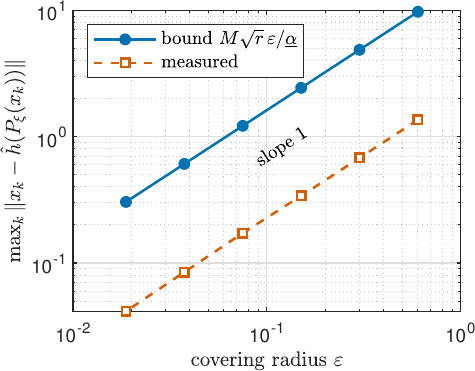}
	\end{minipage}\hfill
	\begin{minipage}[t]{0.38\textwidth}\centering
		\textbf{B}\\[2pt]\includegraphics[width=\linewidth]{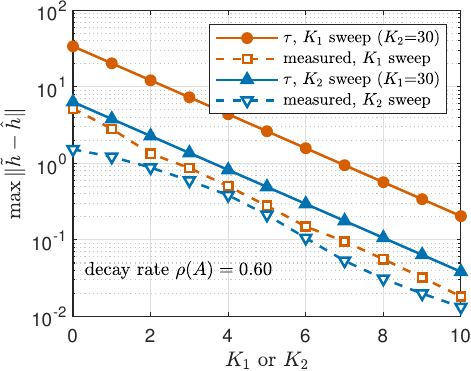}
	\end{minipage}\\[6pt]
	\begin{minipage}[t]{0.38\textwidth}\centering
		\textbf{C}\\[2pt]\includegraphics[width=\linewidth]{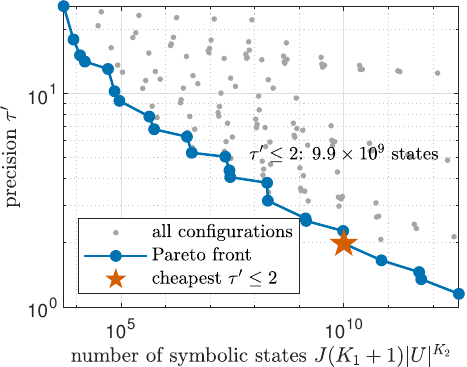}
	\end{minipage}\hfill
	\begin{minipage}[t]{0.38\textwidth}\centering
		\textbf{D}\\[2pt]\includegraphics[width=\linewidth]{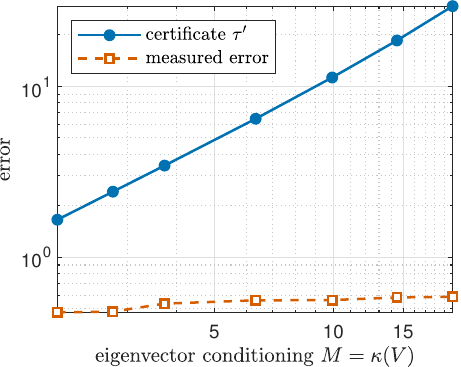}
	\end{minipage}
	\caption{\textbf{Quantitative validation of accuracy, complexity, and
		conservatism on the three-dimensional example.}
		(\textbf{A})~Partition error $\|x_k-\hat{h}(P_{\xi}(x_k))\|$, maximized over
		$4000$ random trajectories of $21$ steps, against the bound of
		Lemma~\ref{lemma_celldiam} as the covering radius $\varepsilon$ is varied;
		both scale linearly in $\varepsilon$ (slope $1$ on the log--log axes).
		(\textbf{B})~Truncation error $\|\tilde{h}-\hat{h}\|$, maximized over $4000$
		random input strings of $45$ steps, against the precision $\tau$ of
		Theorem~\ref{theorem_approx}: the $K_1$ sweep holds $K_2=30$ fixed, the
		$K_2$ sweep holds $K_1=30$ fixed, and all curves decay at the geometric rate
		$\rho(A)$.
		(\textbf{C})~Precision--complexity trade-off over
		$\varepsilon\in\{0.6\cdot2^{-j}\}_{j=0}^{5}$, $K_1\in\{2,4,6,8\}$,
		$K_2\in\{1,\ldots,6\}$: gray dots are individual configurations, the solid
		curve the Pareto front, the star the cheapest configuration certifying
		$\tau'\le2$ ($9.9\times10^{9}$ symbolic states).
		(\textbf{D})~Increasing the eigenvector conditioning $M=\kappa(W)$ from
		$2.0$ to $20$ at fixed $\rho(A)$ inflates the certificate $\tau'$ from
		$1.7$ to $29$ while the measured error grows only from $0.48$ to $0.59$:
		the conditioning governs the conservatism of the certificate, not the
		abstraction itself.
		Parameters: real eigenvalue $\mu_1=0.6$ and complex pair
		$\mu_{2,3}=0.5e^{\pm\mathrm{i}\pi/4}$; $\mathcal{I}=[-1,1]^3$,
		$U=\{\mathbf{0},\pm0.4e_i\}$ ($|U|=7$), giving $\rho(A)=0.60$, $M=6.34$,
		$\underline{\alpha}=0.55$ and $\operatorname{cond}(\Pi)=6.85$.}
	\label{fig:approx}
\end{figure}

\begin{figure}
	\centering
	\textbf{A}\\[2pt]\includegraphics[width=0.92\textwidth]{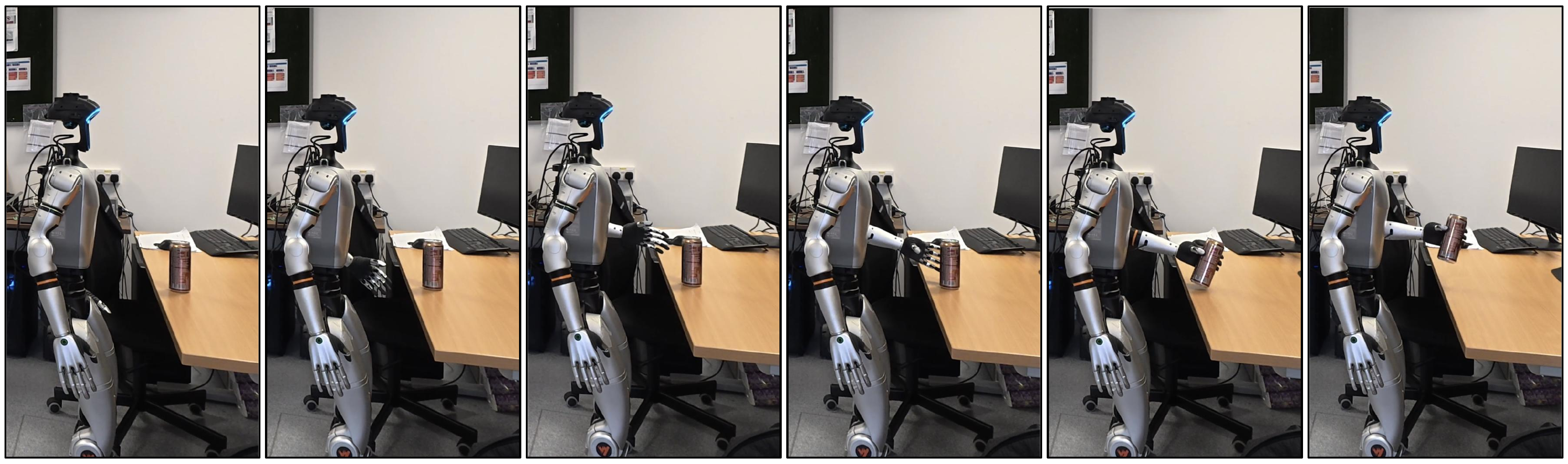}\\[6pt]
	\textbf{B}\\[2pt]\includegraphics[width=0.92\textwidth]{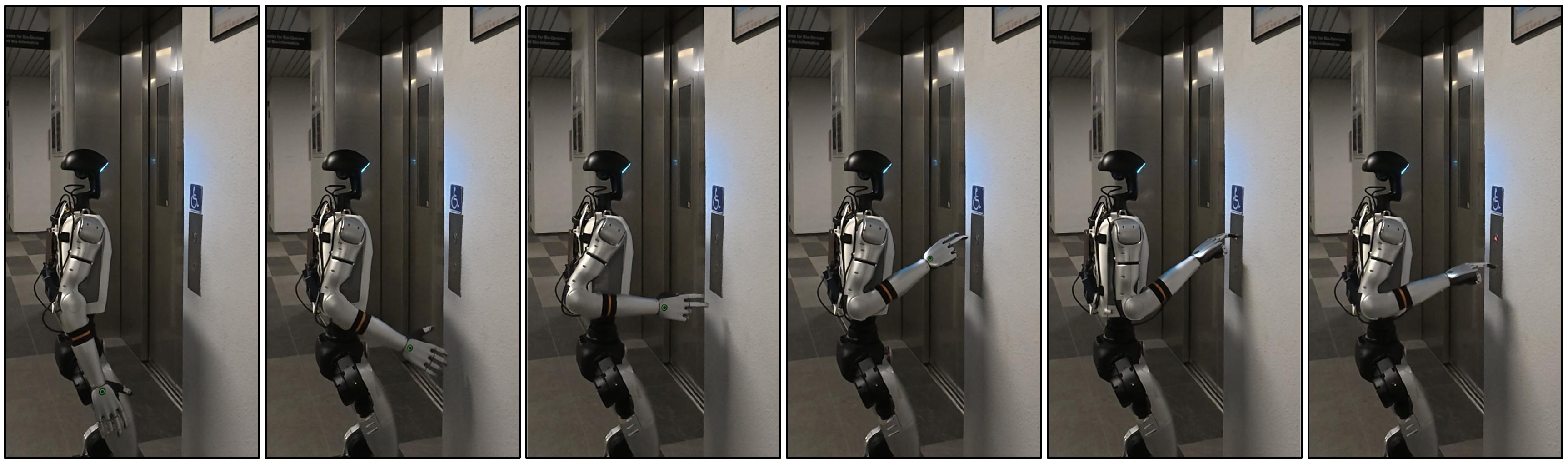}\\[6pt]
	\textbf{C}\\[2pt]\includegraphics[width=0.92\textwidth]{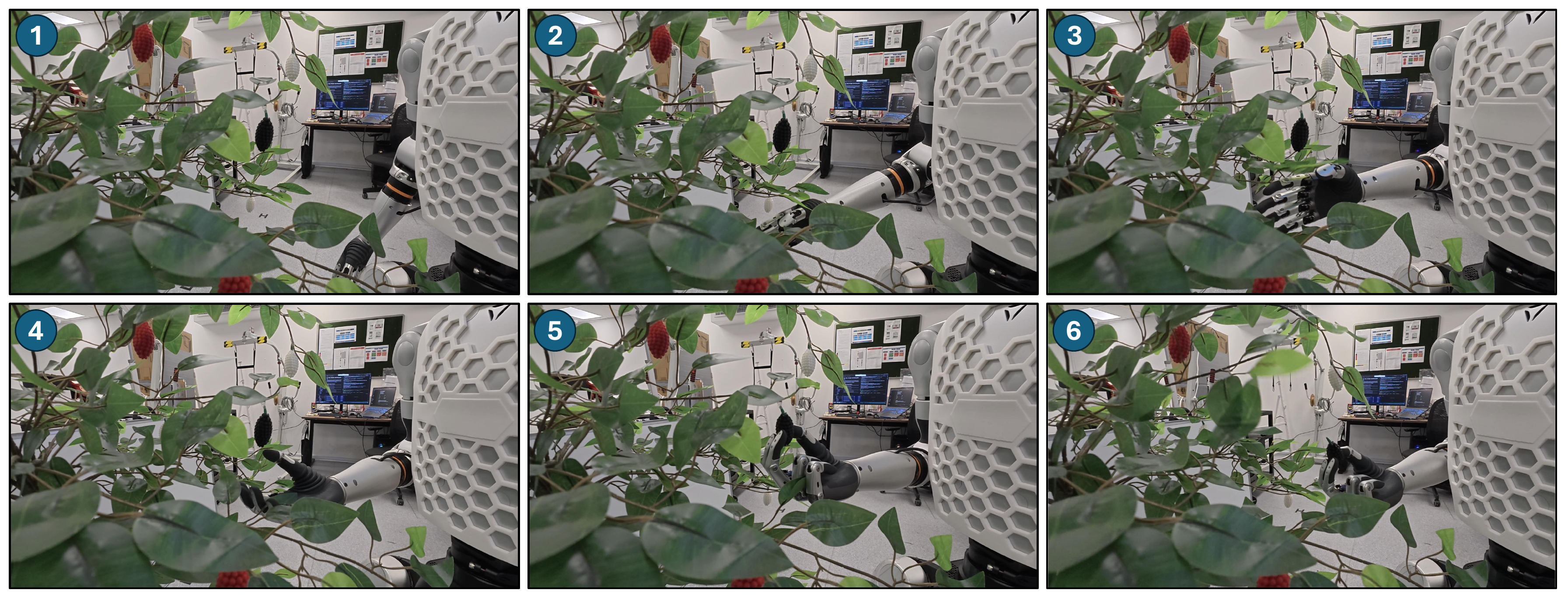}
	\caption{\textbf{The three missions on the Unitree G1 U1 humanoid robot.}
		(\textbf{A})~Object grasping.
		(\textbf{B})~Elevator-button pressing.
		(\textbf{C})~Berry harvesting.
		One symbolic model, built once from the robot's dynamics, serves all three missions;
		only the specification and the labeling of obstacle and target regions change, and
		every certified plan is executed through the platform's existing arm-stretching and
		finger commands.}
	\label{fig:g1_results}
\end{figure}

\clearpage

\bibliography{aipsamp}

\section*{Acknowledgments}
\paragraph*{Funding:}
This work was  jointly supported by the Ministry of Education, Singapore, under AcRF Tier 1 RG164/25, the EEE Ignition Emerging Research Programme 2024 under the project ``Universal Interoperability for Robots as a Service'' and the Robotics Innovation Research Fund (Donation). 
\paragraph*{Author contributions:}
M.L. developed the theory, implemented the software, and performed the simulations and
robot experiments. R.S. conceived the study, formulated the problem of model reuse across
missions, acquired the funding, supervised the research, and revised the manuscript. L.H.
designed the three robot missions and integrated and deployed the planning pipeline on the
Unitree G1 U1. Q.W. built the perception pipeline, performed the experiments, and collected
the experimental data. R.T. analyzed the results and revised the manuscript. All authors
discussed the results and approved the final version.
\paragraph*{Competing interests:}
The authors declare no competing interests.
\paragraph*{Data and materials availability:}
{\sloppy
All data needed to evaluate the conclusions in the paper are present in the paper or the
Supplementary Materials or have been deposited in the repository
\url{https://github.com/mattewkang/Abstraction_experiment}. The precomputed abstraction
maps and grasp-pose networks used on the robot are archived at
\url{https://huggingface.co/Mattewkang/Abstraction_experiment-arm-models}.\par}

\subsection*{Supplementary materials}
Supplementary Text\\
Figs. S1 and S2\\
Movie S1 (the three missions on the Unitree G1 U1)\\
Data S1 (tabulated data underlying Figs.~\ref{fig1} and \ref{fig:approx})

\newpage

\renewcommand{\theequation}{S\arabic{equation}}
\renewcommand{\thefigure}{S\arabic{figure}}
\renewcommand{\thepage}{S\arabic{page}}
\setcounter{equation}{0}
\setcounter{figure}{0}
\setcounter{page}{1}

\begin{center}
{\large\bfseries Supplementary Materials for\\[2pt] \scititle}\\[8pt]
Mingkang~Long, Rong~Su$^{\ast}$, Luhang~Hong, Qingqing~Wang, Ruochen~Tai\\
\small$^\ast$Corresponding author. Email: rsu@ntu.edu.sg
\end{center}

\subsubsection*{This PDF file includes:}
Supplementary Text\\
Figs. S1 and S2\\
Caption for Movie S1\\
Caption for Data S1

\subsubsection*{Other Supplementary Materials for this manuscript:}
Movie S1\\
Data S1

\begin{figure}[H]
	\centering
	\includegraphics[width=0.5\textwidth]{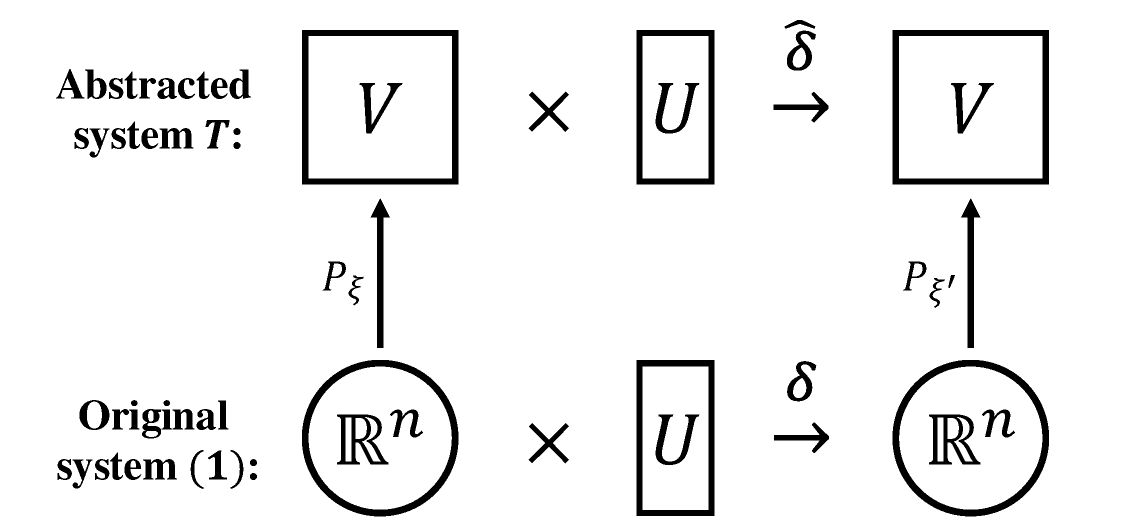}
	\caption{\textbf{Commutative diagram of the abstraction.}
		Relationship between the linear system (Eq.~\ref{eq01}) and the abstracted
		transition system $T$ through the projection $P_{\xi}$.}
	\label{projection}
\end{figure}

\begin{figure}[H]
	\centering
	\includegraphics[width=0.7\textwidth]{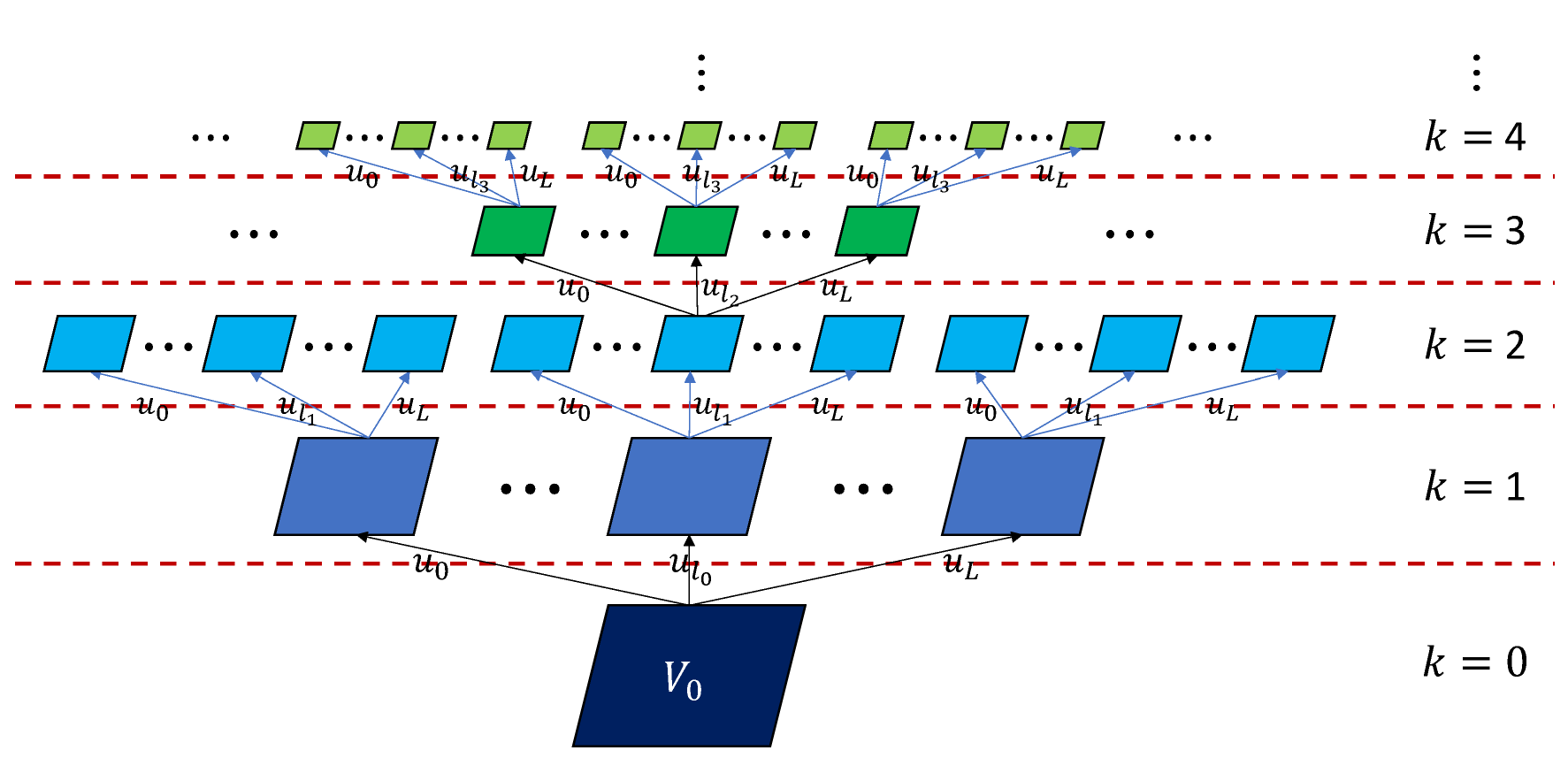}
	\caption{\textbf{Tree-structured abstracted transition system.}
		Each branch represents a transition for a specific input sequence, and each box a
		layer of representative points of the same cardinality as the initial set $V_0$.}
	\label{tree}
\end{figure}

\subsection*{Supplementary Text}
This section collects the proofs of the results stated in Materials and Methods,
together with supporting technical remarks. Equation, lemma and theorem numbers refer
to the main text unless prefixed by~S.

\subsubsection*{Proof of Lemma~\ref{lemma_modal}}
The rows of $\Pi$ are the real and imaginary parts of a basis of left eigenvectors of
the real matrix $A$, so $\Pi$ is the real modal transformation associated with that
basis and is invertible. Separating the real and imaginary parts of $c_i^HA=\mu_ic_i^H$
gives $a_i^TA=\gamma_ia_i^T-\omega_ib_i^T$ and $b_i^TA=\omega_ia_i^T+\gamma_ib_i^T$,
which is the first identity in (Eq.~\ref{eq_blockscaling}); the real case is immediate.
Since $\Lambda_i^T\Lambda_i=(\gamma_i^2+\omega_i^2)I=|\mu_i|^2I$, $\Lambda_i$ scales
every vector by $|\mu_i|$. Finally,
$\|\Pi z\|^2=\sum_{i=1}^{r}\|\pi_i(z)\|^2=\sum_{i=1}^{r}|\langle c_i,z\rangle|^2\geq\underline{\alpha}^2\|z\|^2$
with $\underline{\alpha}>0$, which yields (Eq.~\ref{eq_normequiv}). $\blacksquare$

\begin{remark}
   \label{rmk_norm}
   Each $\dc{i}{\cdot}{\cdot}$ vanishes on $\ker(a_i^T)\cap\ker(b_i^T)$, of dimension
   $n-2$ for $\mu_i\notin\mathbb{R}$ and $n-1$ for $\mu_i\in\mathbb{R}$, so a single
   eigen-direction separates points only when $n=2$ and $\mu_i\notin\mathbb{R}$.
   Inequality (Eq.~\ref{eq_normequiv}) removes this degeneracy by aggregation over $C$,
   making $\max_i\dc{i}{\cdot}{\cdot}$ a norm equivalent to the Euclidean one with
   constant $1/\underline{\alpha}$; this equivalence underlies all the accuracy
   estimates.
\end{remark}

\begin{remark}
   \label{rmk_diag}
   Assumption~\ref{assumption_diag} is exact rather than a convenience: the left
   eigenvectors of $A$ span $\mathbb{C}^n$ iff $A$ is diagonalizable, which is exactly
   the case in which $\max_i\dc{i}{\cdot}{\cdot}$ is a norm. A defective $A$ has fewer than $n$
   independent left eigenvectors, so the cells extend unboundedly along the remaining
   subspace; for the sampled double integrator
   $A=\left[\begin{smallmatrix}1&\Delta t\\ 0&1\end{smallmatrix}\right]$ the only left
   eigenvector is $c^H=[0,~1]$, and positions cannot be separated. When $(A,B)$ is
   controllable an inner feedback loop $u=-Kx+w$ assigns $n$ distinct closed-loop poles,
   making $A_{cl}=A-BK$ diagonalizable, so diagonalizability is a mild requirement on the
   designer's inner loop, whose conditioning $\underline{\alpha}$ also governs accuracy.
   Nonsingularity is equally mild: a sampled-data model $A=e^{A_c\Delta t}$ is always
   nonsingular, and the same pole assignment can place all closed-loop poles away from
   the origin; a singular $A$ would instead collapse distinct representatives under
   propagation, which index preservation must exclude.
\end{remark}

\subsubsection*{Proof of Lemma~\ref{lemma1}}
As $\Pi$ is continuous and $\mathcal{I}$ compact, $\Pi(\mathcal{I})$ is contained in a
bounded box $\prod_{i=1}^{r}B_i$. For each $i$ let $G_i$ be a uniform grid covering
$B_i$ with spacing $2\varepsilon/\sqrt{d_i}$; the distance from any point of $B_i$ to
the nearest grid point is at most $\tfrac{1}{2}\sqrt{d_i}\cdot 2\varepsilon/\sqrt{d_i}=\varepsilon$.
Set $\mathcal{I}_A=\Pi^{-1}(G_1\times\cdots\times G_r)$, which is finite and
eigen-separable. Given $x\in\mathcal{I}$, let $g_i\in G_i$ be nearest to $\pi_i(x)$ and
put $v_0=\Pi^{-1}(g_1,\ldots,g_r)$. Then $\dc{i}{x}{v_0}=\|\pi_i(x)-g_i\|\leq\varepsilon$
for every $i$, which is (Eq.~\ref{eq_cover_multi}); applying (Eq.~\ref{eq_normequiv})
to $z=x-v_0$ gives $\|x-v_0\|\leq\sqrt{r}\,\varepsilon/\underline{\alpha}$.
$\blacksquare$

\begin{remark}
   \label{rmk_constructive}
   The proof is constructive: it yields an explicit eigen-separable seed set
   $\mathcal{I}_A$ and the bound $\sqrt{r}\,\varepsilon/\underline{\alpha}$ on the
   Euclidean distance from any initial state to its seed, known before the abstraction
   is built. Both are prerequisites for the quantitative accuracy analysis, in which
   $\varepsilon$ enters the precision estimate (Eq.~\ref{eq_tauprime}) directly through
   the term $M\sqrt{r}\,\varepsilon/\underline{\alpha}$.
\end{remark}

\begin{remark}
   \label{rmk_product}
   By Definition~\ref{definition2}, $N(v)=\bigcap_{i=1}^{r}N_{c_i}(v)$, where
   \[N_{c_i}(v)=\big\{x\in\mathbb{R}^n\mid\dc{i}{x}{v}\leq\dc{i}{x}{v'}~\forall v'\in\mathbb{V}\big\}\]
   is the one-directional cell of $v$ along $c_i$, so $N(v)$ is the common
   refinement of the $r$ one-directional partitions. Since $\dc{i}{x}{v'}$ depends on
   $v'$ only through $\pi_i(v')\in G_i$, the $r$ minimizations decouple and, after the
   coordinatewise tie-break on each $G_i$, are jointly attained by the unique
   $v=\Pi^{-1}(g_1,\ldots,g_r)\in\mathbb{V}$. Hence the cells cover $\mathbb{R}^n$, are
   pairwise disjoint, and each contains its representative.
\end{remark}

\subsubsection*{Proof of Theorem~\ref{theorem1}}
\textit{Step 1 (each layer is eigen-separable).} By (Eq.~\ref{eq04}), we have that
$\hat{h}(v_k,\xi)=A^kv_0+w_{\xi}$ for some $v_0\in\mathcal{I}_A$ with $w_{\xi}=\sum_{i=0}^{k-1}A^iBu_{l_{k-1-i}}$. So by (Eq.~\ref{eq_blockscaling})
$\pi_i(\hat{h}(v_k,\xi))=\Lambda_i^{k}\pi_i(v_0)+\pi_i(w_{\xi})$ and
\begin{equation}
   \label{eq_layerproduct}
   \Pi\big(\hat{h}(\tilde{V})\big)=\prod_{i=1}^{r}\Big(\Lambda_i^{k}G_i+\pi_i(w_{\xi})\Big),
\end{equation}
again a product set; since $A$ is nonsingular (Assumption~\ref{assumption_diag}), each
$\Lambda_i$ is invertible, so $\Lambda_i^{k}G_i$ retains $|G_i|$ distinct points and
every layer inherits the eigen-separability of $\mathcal{I}_A$. By
Remark~\ref{rmk_product}, the $r$ minimizations in (Eq.~\ref{eq_proj}) are then
attained by a single $v$; hence $P_{\xi}$ is well defined.

\textit{Step 2 (one-step index preservation).} Let $v=(v_k,\xi)$ and $u\in U$; with
$x^{+}=Ax+Bu$ and $v^{+}=\hat{\delta}(v,u)$, the recursion (Eq.~\ref{eq_hrec}) gives
$\hat{h}(v^{+})=A\hat{h}(v)+Bu$, so for every $i$,
\begin{equation}
   \label{eq_contract}
   \dc{i}{x^{+}}{\hat{h}(v^{+})}=\big|\big\langle c_i,\ A(x-\hat{h}(v))\big\rangle\big|=|\mu_i|\,\dc{i}{x}{\hat{h}(v)},
\end{equation}
using $c_i^HA=\mu_ic_i^H$. If $P_{\xi}(x)=(v_k^F,\xi)$, applying (Eq.~\ref{eq_contract})
to every $j$ multiplies all $J$ distances in the $i$-th eigen-coordinate by the same
factor $|\mu_i|$, which is positive because $A$ is nonsingular
(Assumption~\ref{assumption_diag}). Scaling by a positive constant preserves each
per-coordinate arg-min set together with its set of tied points, and the order on
$\Lambda_i^{k+1}G_i$ is the image of the order on $\Lambda_i^{k}G_i$, so the
coordinatewise tie-break selects the image of the previously selected point and the
minimizer is unchanged; as this holds for
each $i$ and the joint minimizer is determined blockwise, (Eq.~\ref{eq_indexpres})
follows.

\textit{Step 3 (bisimulation).} Let
$P_{\xi}(x)=P_{\xi}(y)=(v_k^F,\xi)$ for some index $F$, which is condition~(1) of
Definition~\ref{definition4}. Applying (Eq.~\ref{eq_indexpres})
inductively along any $\eta\in U^{*}$ gives
$P_{\xi\eta}(\delta(x,\eta))=P_{\xi\eta}(\delta(y,\eta))=(v_{k+|\eta|}^F,\xi\eta)$,
that is, $\delta(x,\eta)\equiv_{\xi\eta}\delta(y,\eta)$; since the transition relation
is total on $U$, $\delta(x,\eta)!$ and $\delta(y,\eta)!$ for every $\eta$, so
condition~(2) of Definition~\ref{definition4} holds as well. $\blacksquare$

\begin{remark}
   \label{rmk_boundary}
   A boundary point, where $\pi_i(x)$ is equidistant from two representatives in some
   direction, is resolved deterministically by the coordinatewise tie-break of Definition~\ref{definition2}, so every $x$ is
   assigned to exactly one half-open cell. By (Eq.~\ref{eq_contract}) every projection
   distance is scaled by the same factor $|\mu_i|>0$ (Assumption~\ref{assumption_diag})
   under the dynamics, so a tie at one step remains a tie at the next and the cell
   assignment of a boundary point is preserved along the whole trajectory, as for
   interior points.
\end{remark}

\begin{remark}
   \label{rmk_recover}
   Retaining a single eigen-direction ($r=1$) is admissible and yields the coarsest,
   still exactly bisimilar member of the family; additional directions improve metric
   accuracy (Lemma~\ref{lemma_celldiam}), the cells becoming arbitrarily small as
   $\varepsilon\to0$, at the price of $J=\prod_{i=1}^{r}|G_i|$ growing as
   $\varepsilon^{-n}$ instead of $\varepsilon^{-d_1}$.
\end{remark}

\subsubsection*{Proof of Corollary~\ref{corollary1}}
Let $x\in\mathbb{R}^n$, $\xi\in U^{*}$ and $v=P_{\xi}(x)=(v_k^F,\xi)$ with $k=|\xi|$.
Whenever $\delta(x,\eta)!$, applying the index-preservation property
(Eq.~\ref{eq_indexpres}) inductively along $\eta$ gives
$P_{\xi\eta}(\delta(x,\eta))=(v_{k+|\eta|}^F,\xi\eta)=\hat{\delta}(v,\eta)$,
where the last equality is (Eq.~\ref{eq03}) iterated along $\eta$. $\blacksquare$

\begin{remark}
   \label{rmk_onto}
   The correspondence of Corollary~\ref{corollary1} is onto. Let
   $v=(v_0,\epsilon)\in V_0$ and $\xi\in U^{*}$ with $\hat{\delta}(v,\xi)!$, and take
   $x=v_0$, the representative point of the cell $N(v_0)$. By
   Remark~\ref{rmk_product} every cell contains its own representative, so
   $P_{\epsilon}(x)=v$. Comparing (Eq.~\ref{eq01}) with the recursion (Eq.~\ref{eq_hrec})
   shows that the trajectory of the representative point coincides with the propagated
   representative, $\delta(x,\xi)=\hat{h}(\hat{\delta}(v,\xi))$; hence every minimum in
   (Eq.~\ref{eq_proj}) is attained at distance zero by $\hat{\delta}(v,\xi)$ itself.
   Since $A$ is nonsingular (Assumption~\ref{assumption_diag}), distinct initial
   representatives propagate to distinct layer representatives,
   \[\hat{h}\big((v_{|\xi|}^{j_1},\xi)\big)-\hat{h}\big((v_{|\xi|}^{j_2},\xi)\big)=A^{|\xi|}\big(v_0^{j_1}-v_0^{j_2}\big)\neq\mathbf{0}\]
   for $j_1\neq j_2$, and $\Pi$ is invertible, so no other layer state attains distance
   zero in every eigen-coordinate; the distance-zero minimizer is therefore unique and
   $P_{\xi}(\delta(x,\xi))=\hat{\delta}(v,\xi)$, i.e. every path of $T$ is the projection
   of a trajectory of $\Sigma$.
\end{remark}

\subsubsection*{Proof of Theorem~\ref{theorem2}}
Define $\Psi^{\prime}:\mathbb{R}^n\times U^*\rightarrow\{0,1\}$ by
\begin{equation}
   \label{eq_minhull}
   \Psi^{\prime}(x,\xi)=1\iff(\forall y\in\mathbb{R}^n)\big[P_{\epsilon}(y)=P_{\epsilon}(x)\Rightarrow\Psi(y,\xi)=1\big].
\end{equation}
\textit{Block conservative.} The right-hand side of (Eq.~\ref{eq_minhull}) depends on $x$
only through $P_{\epsilon}(x)$, so $\Psi^{\prime}(x,\xi)=1$ and $P_{\epsilon}(x)=P_{\epsilon}(y)$
imply $\Psi^{\prime}(y,\xi)=1$.
\textit{Conservative hull.} Taking $y=x$ in (Eq.~\ref{eq_minhull}) gives
$\Psi^{\prime}(x,\xi)=1\Rightarrow\Psi(x,\xi)=1$.
\textit{Minimum.} Let $\Psi^{\prime\prime}$ be any conservative hull of $\Psi$ and
$\Psi^{\prime\prime}(x,\xi)=1$. For every $y$ with $P_{\epsilon}(y)=P_{\epsilon}(x)$, block
conservativeness of $\Psi^{\prime\prime}$ gives $\Psi^{\prime\prime}(y,\xi)=1$, hence
$\Psi(y,\xi)=1$; by (Eq.~\ref{eq_minhull}), $\Psi^{\prime}(x,\xi)=1$. If
$\Psi^{\prime\prime}$ is also minimum, the same implication holds in both directions, so
$\Psi^{\prime\prime}=\Psi^{\prime}$ and the minimum hull is unique.
\textit{Extension from the representative.} Let $(v_0,\epsilon)\in V_0$. By
Remark~\ref{rmk_product} the representative point lies in its own cell,
$P_{\epsilon}(v_0)=(v_0,\epsilon)$, so every $x\in P_{\epsilon}^{-1}((v_0,\epsilon))$
satisfies $P_{\epsilon}(x)=P_{\epsilon}(v_0)$, and block conservativeness yields
$\Psi^{\prime}(v_0,\xi)=1\Rightarrow\Psi^{\prime}(x,\xi)=1$. $\blacksquare$

\subsubsection*{Proof of Theorem~\ref{theorem_approx}}
\textit{Step 1 (observation gap between $S$ and $T$).} The two systems share the input
alphabet, the representative points and the transition structure. Unrolling (Eq.~\ref{eq_hrec}),
the observation of $T$ is $\hat{h}=A^kv_0+\sum_{i=0}^{k-1}A^iBu_{l_{k-1-i}}$, while by
(Eq.~\ref{eq_htilde}) that of $S$ is
$\tilde{h}=v_k+\sum_{i=0}^{K_2-1}A^iBu_{l_{k-1-i}}$, so
\begin{equation}
   \label{eq_gapdecomp}
   \tilde{h}(\cdot)-\hat{h}(\cdot)=\underbrace{\big(v_k-A^kv_0\big)}_{\text{state truncation}}-\underbrace{\sum_{i=K_2}^{k-1}A^iBu_{l_{k-1-i}}}_{\text{input truncation}}.
\end{equation}
For the state term, the difference vanishes for $k\leq K_1$; for $k>K_1$,
$\|v_k-A^kv_0\|=\|A^kv_0\|\leq M r_{\mathcal{I}}\rho(A)^{k}\leq M r_{\mathcal{I}}\rho(A)^{K_1+1}$
by (Eq.~\ref{eq_powerbound}). For the input term,
\[\big\|\sum_{i=K_2}^{k-1}A^iBu_{l_{k-1-i}}\big\|\leq M r_U\sum_{i=K_2}^{\infty}\rho(A)^{i}=M r_U\rho(A)^{K_2}/(1-\rho(A)).\]
The triangle inequality gives
\begin{equation}
   \label{eq_seedgap}
   \big\|\tilde{h}(q)-\hat{h}(v)\big\|\leq\tau
\end{equation}
for every state $q$ of $S$ and state $v$ of $T$ matched by the same index $j$ and the same
input history; the bound is uniform in $k$, the two systems accept the same input strings,
and the index $j$ is preserved along every input string, so the successors under any common
input string satisfy the same bound.

\textit{Step 2 (approximate bisimilarity).} Let $\tilde{P}_{\xi}(x)=\tilde{P}_{\xi}(y)$, which
is condition~(1) of Definition~\ref{definition5}. Since the transition relation of
(Eq.~\ref{eq01}) is total on $U$, $\delta(x,\eta)!$ if and only if $\delta(y,\eta)!$ for
every $\eta\in U^*$. By Step~1, the seeds $\tilde{h}(\cdot)$ of the layer of $S$ reached by
$\xi\eta$ differ from the exact seeds $\hat{h}(\cdot)$ of the corresponding layer of $T$ by
at most $\tau$, uniformly in $\eta$, while the exact seeds are transported by the dynamics
with index preservation (Eq.~\ref{eq_indexpres}); hence, whenever
$\tilde{P}_{\xi\eta}(\delta(x,\eta))\neq\tilde{P}_{\xi\eta}(\delta(y,\eta))$, the two
successors are separated only by the $\tau$-displacement of the cell boundaries of $S$
relative to those of $T$, which gives condition~(2) with the approximation error $\tau$ of
(Eq.~\ref{eq_tau}). $\blacksquare$

\begin{remark}
   \label{rmk_Mconditioning}
   The constant $M$ in (Eq.~\ref{eq_powerbound}) measures how far the eigenvector basis
   of $A$ is from orthogonal, the same kind of quantity as $\underline{\alpha}=\sigma_{\min}(\Pi)$;
   both degrade as $A$ approaches a defective matrix. The conditioning of the
   eigen-decomposition thus governs both error sources, a reason to report
   $\operatorname{cond}(\Pi)$ and prefer well-separated closed-loop poles.
\end{remark}

\subsubsection*{Proof of Lemma~\ref{lemma_celldiam}}
Let $P_{\epsilon}(x_0)=(v_0^F,\epsilon)$. By index preservation (Eq.~\ref{eq_indexpres}),
$P_{\xi}(x_k)=(v_k^F,\xi)$. Since $x_k$ and $\hat{h}(v_k^F,\xi)$ are propagated from
$x_0$ and $v_0^F$ under the same input string by (Eq.~\ref{eq01}) and (Eq.~\ref{eq_hrec}),
subtracting the recursions gives
\begin{equation}
   \label{eq_gapprop}
   x_k-\hat{h}\big(P_{\xi}(x_k)\big)=A^{k}\big(x_0-v_0^F\big),
\end{equation}
and Lemma~\ref{lemma1} bounds $\|x_0-v_0^F\|$ by $\sqrt{r}\,\varepsilon/\underline{\alpha}$.
$\blacksquare$

\subsubsection*{Proof of Theorem~\ref{theorem3}}
Let $k=|\xi|$, $x_k=\delta(x_0,\xi)$, $v=\hat{\delta}((v_0^F,\epsilon),\xi)=(v_k^F,\xi)$ the state of $T$
reached from the initial cell of $x_0$, and $q=\tilde{\delta}((v_0^F,\epsilon),\xi)$ the state of $S$
reached from the same initial state under the same input history. By Corollary~\ref{corollary1},
$P_{\xi}(x_k)=v$, so Lemma~\ref{lemma_celldiam} together with (Eq.~\ref{eq_powerbound}) gives
\[\|x_k-\hat{h}(v)\|\leq\|A^k\|\,\frac{\sqrt{r}\,\varepsilon}{\underline{\alpha}}\leq\frac{M\sqrt{r}\,\varepsilon}{\underline{\alpha}}.\]
The states $v$ and $q$ carry the same index $F$ and the same input history, so (Eq.~\ref{eq_seedgap})
in the proof of Theorem~\ref{theorem_approx} gives $\|\tilde{h}(q)-\hat{h}(v)\|\leq\tau$. The triangle
inequality
\[\|x_k-\tilde{h}(q)\|\leq\|x_k-\hat{h}(v)\|+\|\hat{h}(v)-\tilde{h}(q)\|\leq\frac{M\sqrt{r}\,\varepsilon}{\underline{\alpha}}+\tau=\tau^{\prime}\]
is (Eq.~\ref{eq_tauprime_bound}). The bound is uniform in $k$ because both terms are. $\blacksquare$

\begin{remark}
   \label{rmk_twoerrors}
   Partitioning along the full set of eigen-directions is indispensable, since a single
   direction bounds only a seminorm and leaves the abstraction term unbounded for
   $n\geq3$, which no horizon can compensate. Unlike~\cite{Girard2007Approximately}, the
   construction additionally specifies through $P_{\xi}$ how to assign a representative
   point to any continuous state, and reduces to an exactly bisimilar finite abstraction
   over a finite horizon.
\end{remark}

\clearpage

\paragraph{Caption for Movie S1.}
\textbf{The three missions of Fig.~\ref{fig:g1_results} executed on the Unitree G1 U1
humanoid robot with one abstracted model.} The video shows, in the order of the main
text, object grasping on a cluttered tabletop (Fig.~\ref{fig:g1_results}A),
elevator-button pressing (Fig.~\ref{fig:g1_results}B) and berry harvesting among
foliage (Fig.~\ref{fig:g1_results}C). All footage is played in real time. In each
mission the robot starts from its home configuration, the certified joint-space path
synthesized on the relabeled abstraction is executed as a sequence of the vendor's
arm-stretching commands, and the terminal transition into the \textsf{goal} cell is
followed by the finger command (closing on the object, pressing the button, or closing
on the berry). The same abstracted model, joint-space quantization and planning pipeline
are used throughout; only the specification and the labeling of obstacle and target
cells differ between the three missions.

\paragraph{Caption for Data S1.}
\textbf{Tabulated data underlying Figs.~\ref{fig1} and \ref{fig:approx} and the
index-preservation statistics quoted in the text.} The archive \texttt{data\_S1.zip}
contains one CSV file per table; the file \texttt{README.csv} lists every table with the
figure panel it corresponds to. Tables \texttt{Fig3AD}, \texttt{Fig3BE} and \texttt{Fig3CF}
give, for each column of Fig.~\ref{fig1}, the seed coordinates (\texttt{\_seeds}) and
the sampled states with their cell index (\texttt{\_states}) at $k=0$ and after the
common input $u$ at $k=1$; \texttt{Fig3\_systems} lists the three system matrices and
their eigenvalues. Tables \texttt{Fig4A\_partition} to \texttt{Fig4D\_conditioning}
give the swept parameter, the measured maximum error and the analytical bound for each
panel of Fig.~\ref{fig:approx}; \texttt{Fig4C\_tradeoff} lists all $144$
$(\varepsilon,K_1,K_2)$ configurations with their number of symbolic states and
precision $\tau'$, flagging the Pareto front and the cheapest configuration certifying
$\tau'\le2$. \texttt{System\_parameters} records $\rho(A)$, $M$, $\underline{\alpha}$,
$\operatorname{cond}(\Pi)$, $r$, $|U|$ and the covering radius, and
\texttt{Text\_index\_preservation} records the $5\times10^{5}$ one-step checks of
Eq.~\ref{eq_indexpres}. All values are generated by the MATLAB scripts
\texttt{sr\_error\_figs.m} (\texttt{rng(0)}) and \texttt{neighborhood\_figs.m}
(\texttt{rng(1)}) in the repository cited in the Acknowledgments.

\end{document}